\documentclass[12pt,a4paper]{article}

\usepackage{graphicx}
\usepackage{amssymb}
\usepackage{amsmath}
\usepackage{amsfonts}
\usepackage{listings}
\usepackage{algorithm}
\usepackage{algpseudocode}
\usepackage{enumitem}
\usepackage{verbatim}
\usepackage{amsthm}
\usepackage{array}
\usepackage[utf8]{inputenc}
\usepackage{caption}
\usepackage{subcaption}
\usepackage{color}
\usepackage{url}

\newcommand{\bx}{\mathbf{x}}

\newcommand{\bz}{\mathbf{z}}

\newcommand{\beps}{\boldsymbol{\epsilon}}

\newcommand{\bepstheta}{\boldsymbol{\epsilon}_\theta}

\newcommand{\bzero}{\mathbf{0}}

\newcommand{\oa}{\overline{\alpha}}

\newcommand{\bI}{\mathbf{I}}

\begin{document}

\author{
Lucas Beerens
 \thanks{%
           School of Mathematics,
           University of Edinburgh,
           EH9 3FD, UK.
            Supported by 
            MAC-MIGS Centre for Doctoral Training under EPSRC grant EP/S023291/1.
           (\texttt{l.beerens@sms.ed.ac.uk})
           }
\and
Catherine F. Higham%
  \thanks{%
           School of Computing Science,
           University of Glasgow,
           Sir Alwyn Williams Building, Glasgow, G12 8QQ.
            Supported by EPSRC grant EP/T00097X/1. 
           (\texttt{Catherine.Higham@glasgow.ac.uk})
           }
           \and
        Desmond J. Higham%
            \thanks{%
           School of Mathematics,
           University of Edinburgh,
           EH9 3FD, UK.
            Supported by EPSRC grant EP/V046527/1. 
           (\texttt{d.j.higham@ed.ac.uk})
           }  
        }

\date{}

\title{Deceptive Diffusion: Generating Synthetic Adversarial Examples
}

\maketitle
 \begin{abstract}
 We introduce the concept of deceptive diffusion---training a generative AI model to produce adversarial images. Whereas  
 a traditional adversarial attack algorithm aims to perturb an existing image to induce a misclassificaton, the deceptive diffusion model
 can create an arbitrary number of new, misclassified images that are not directly associated with training or test images.
 Deceptive diffusion offers the possibility of strengthening defence algorithms by providing 
 adversarial training data at scale, including types of misclassification that are otherwise difficult to find.
 In our experiments, we also investigate the effect of training on 
 a partially attacked data set.
 This highlights a new type of vulnerability for generative diffusion models: if an attacker is able to stealthily 
 poison a portion of the training data, then the resulting diffusion model 
 will generate a similar proportion of misleading outputs.
\end{abstract}

\section{Motivation}\label{sec:mot}
In this work, we combine two types of algorithm 
that have come to prominence in artificial intelligence (AI):  
adversarial and generative.
Adversarial attack algorithms
are designed to reveal vulnerabilities in classification systems; for example by perturbing a chosen image in a way that is imperceptible to the human eye, but causes a change in classification 
\cite{goodfellow2014explaining,szegedy2013intriguing}.
Generative models are designed to create outputs that are similar to, but
not simply copies of, the examples on which they were trained 
\cite{Dhar21,Ho20}.
Here, we show that by training on 
data that consists of adversarially perturbed images, 
a generative diffusion model can be made to create fresh examples of 
adversarial images that do not correspond directly to any underlying real images.

In section~\ref{sec:bg} we give some background information on the two main ingredients of our work: adversarial attack algorithms and generative diffusion models.
Section~\ref{sec:exp} describes the results of computational experiments where we investigate the idea of training a diffusion model on 
adversarially-perturbed data.
We finish with a brief discussion in Section~\ref{sec:conc}.

\subsection{Related Work}

We refer to \cite{Chen_2023_ICCV} for an overview of recent attempts to 
use generative AI tools to produce adversarial inputs.
So far, the AdvDiffuser algorithm of
\cite{Chen_2023_ICCV} appears to be the first and only 
approach to generating new, synthesized, examples of adversarial images 
using a diffusion model.
In that work, the authors take an existing, trained diffusion model 
and adapt the denoising, or backward, process by 
 adding adversarial perturbations at each time step.
 This change increases computational complexity, since an extra gradient step is required at each time point.
 Our approach differs by building a new diffusion model, which then
  generates images with a standard de-noising algorithm.
In addition to lowering the computational cost, our \emph{deceptive diffusion} method 
reveals a new type of security threat that arises when standard generative  
diffusion models are created on training data that 
has been attacked.
  In particular, we find that the drop in classification success is in direct proportion to the fraction of training data that is
  adversarially perturbed.
Hence, if an attacker is able to 
poison some portion of the training data, the builders of a generative diffusion model may inadvertently create a tool that produces a corresponding proportion of  
adversarial images.

\section{Background}\label{sec:bg}

\subsection{Adversarial Attack Algorithms}\label{subsec:adv}

State of the art image classification tools are known to possess
inherent vulnerabilites. In particular, they can be fooled by 
adversarial attacks, where an existing image undergoes a small perturbation 
that would not be noticeable to a human, but causes a change in the predicted class.
Since this effect was first pointed out,
\cite{goodfellow2014explaining,szegedy2013intriguing},
a wide range of attack and defence strategies
have been put forward,
\cite{Attack_survey_2018,ink23,Madry18,mmstv18,deepfool},
and bigger picture questions concerning 
the inevitability  of attack success have been investigated,
\cite{Hansen21,fawzi18,shafahi2018adv,thwg21,tyukin2020adversarial}.
The susceptibility of AI systems to attack is a serious issue  
in many application areas and it is pertinent to the recent calls for 
AI regulation. For example, the 
amendment of June 2023 \cite{euaiactamendment} to
Article 15 – paragraph 4 – subparagraph 1 
of the 
EU AI act \cite{euaiact} requires that: 
\emph{``High-risk AI systems shall be resilient as regards to attempts by unauthorised third parties to alter their use, behaviour, outputs or performance by exploiting the system vulnerabilities.''}

\subsection{Generative Diffusion Models}\label{subsec:diff}

A generative diffusion model for creating realistic, but synthetic, images
can be built by first training a neural network to de-noise a collection of noisy images, and then asking the network to de-noise a new sample of pure noise \cite{diff_surv24}.

In Algorithms~\ref{alg:forward} and \ref{alg:backward}
we summarize the basic unconditional diffusion model setting from \cite{Ho20}; see also \cite{HHGdiff,Luo22} for detailed explanations of the steps involved. 
Here, the $\alpha_t$ are parameters taking values between zero and one.
They have the form 
$
\alpha_t = 1 - \beta_t,
$
where the predetermined sequence $\beta_1, \beta_2, \ldots, \beta_{T}$ is  known as the \emph{variance schedule}. 
In \cite{Ho20}, linearly increasing values from $\beta_1 = {10}^{-4}$ to $\beta_T = 0.02$ are used.
We also let 
\[
        \oa_t = \prod_{i=1}^{t} \alpha_i,
        \]
        and
        \[
        \sigma^2_q(t) =  \frac{(1 - \alpha_t) ( 1 - \oa_{t-1}) } { 1 - \oa_t }.
        \]

In step 5 of Algorithm~\ref{alg:forward}
$\bepstheta$ denotes the output from a neural network.
Given a version of the noisy image,
$ \sqrt{\oa_{t}} \, \bx_{0} + \sqrt{ 1 - \oa_{t}} \, \beps$,
corresponding to a time $t$, the job of the network is to 
predict the noise $\beps$.
Here, a simple least-squares loss function is used. 

\begin{algorithm}
\caption{Training with the forward process \cite{Ho20}}\label{alg:forward}
\begin{algorithmic}[1]
\Repeat
\State $\bx_{0}  \sim q(\bx_{0})$  \Comment{choose an image from training set}
\State $t \sim \mathrm{Uniform}(\{1,2,\ldots,T\})$  
\State $\beps \sim \mathrm{N}(\bzero,\bI)$ \Comment{standard Gaussian sample}
\State Take gradient step w.r.t. $\theta$ on $\| \beps - \beps_{\theta}(  \sqrt{\oa_{t}} \, \bx_{0} + \sqrt{ 1 - \oa_{t}} \, \beps, t )\|_2^2$
\Until converged
\end{algorithmic}
\end{algorithm}

Algorithm~\ref{alg:backward} from 
\cite{Ho20} 
summarizes the sampling process.
Here, in step~1 a set of pure noise pixel values is de-noised from time $T$
to time $0$ in order to produce a new synthetic image.

\begin{algorithm}
\caption{Sampling with the backward process \cite{Ho20}}\label{alg:backward}
\begin{algorithmic}[1]
\State $\bx_{T} \sim \mathrm{N}(\bzero,\bI)$ \Comment{standard Gaussian sample} 
\For{$t = T, T-1, \ldots, 1$}
\State $\bz  \sim \mathrm{N}(\bzero,\bI)$ \Comment{standard Gaussian sample} 
\State $\bx_{t-1} = 
\frac{ 1 } {  \sqrt{\alpha_t} }
\left(
\bx_{t}
-
 \frac{ 1 - \alpha_t} {   \sqrt{1-\oa_t} }  \, \bepstheta
 \right)
 + \sigma_q(t) \, \bz$
\EndFor
\State \Return $\bx_{0}$
\end{algorithmic}
\end{algorithm}

\section{Experimental Results}\label{sec:exp}

We now outline the key components in our  
computational experiments.

We use the MNIST data set \cite{lcb-digits_old}, which 
contains 60,000 training images and 10,000 test images of 
handwritten digits, with labels indicating the categories:
`0', `1',`2',\ldots,`9'.

As a classifier, we use a convolutional neural network
based on the architecture of LeNet \cite{LeNet98,LeNet89}.
The exact architecture can be found in our code.
After training, this classifier achieves an accuracy of
99.02\% on the test images.

For the adversarial attack algorithm we use PGDL2 \cite{kim2020torchattacks}, a PyTorch implementation of 
the projected gradient method from \cite{Madry18}.
This attack algorithm uses a robust optimization approach to 
seek an optimal perturbation in an $\ell_2$ sense, using gradients of the loss function.
We use the default setting in PGDL2 where an attack is declared successful 
if it finds a sufficiently small class-changing perturbation within a specified number of iterations of a first order gradient method. 
The bound on the $\ell_2$ norm of the attack was set to 2 (each of the 784 pixels takes values between $0$ and $1$).
We chose a large bound of 1000 on the number of iterations in order 
to maximize the size of the attacked image dataset for the training the diffusion model.
We used PGDL2 in untargeted mode, so that any change of classification is acceptable.

In the diffusion model, we use a neural network with a 
UNet2DModel
architecture from 
\begin{verbatim}
https://huggingface.co/docs/diffusers/en/api/models/unet2d
\end{verbatim}
which is motivated by the original version in 
\cite{Ronn15}.

\subsection{Initial Sanity Check} \label{subsec:sanity}

Before moving on to adversarial images, we first report on an initial test which confirms that the diffusion model is capable of producing 
outputs that are acceptable to the classifier.

In this test, we train the diffusion model using the original MNIST training data. We supply the labels during the training process, so we use 
a conditional version of Algorithm~\ref{alg:forward}, where in step 5 the network
learns to remove noise and produce an image when given both a time $t$ and a label.
This is built in to the UNet2DModel. A trainable encoder maps the label into the same space as the timestep. These two quantities are then added and passed to the model in the same way that the time is usually passed \cite{rom22}.

\begin{figure}
    \centering
     \includegraphics[height = 0.4\textheight]{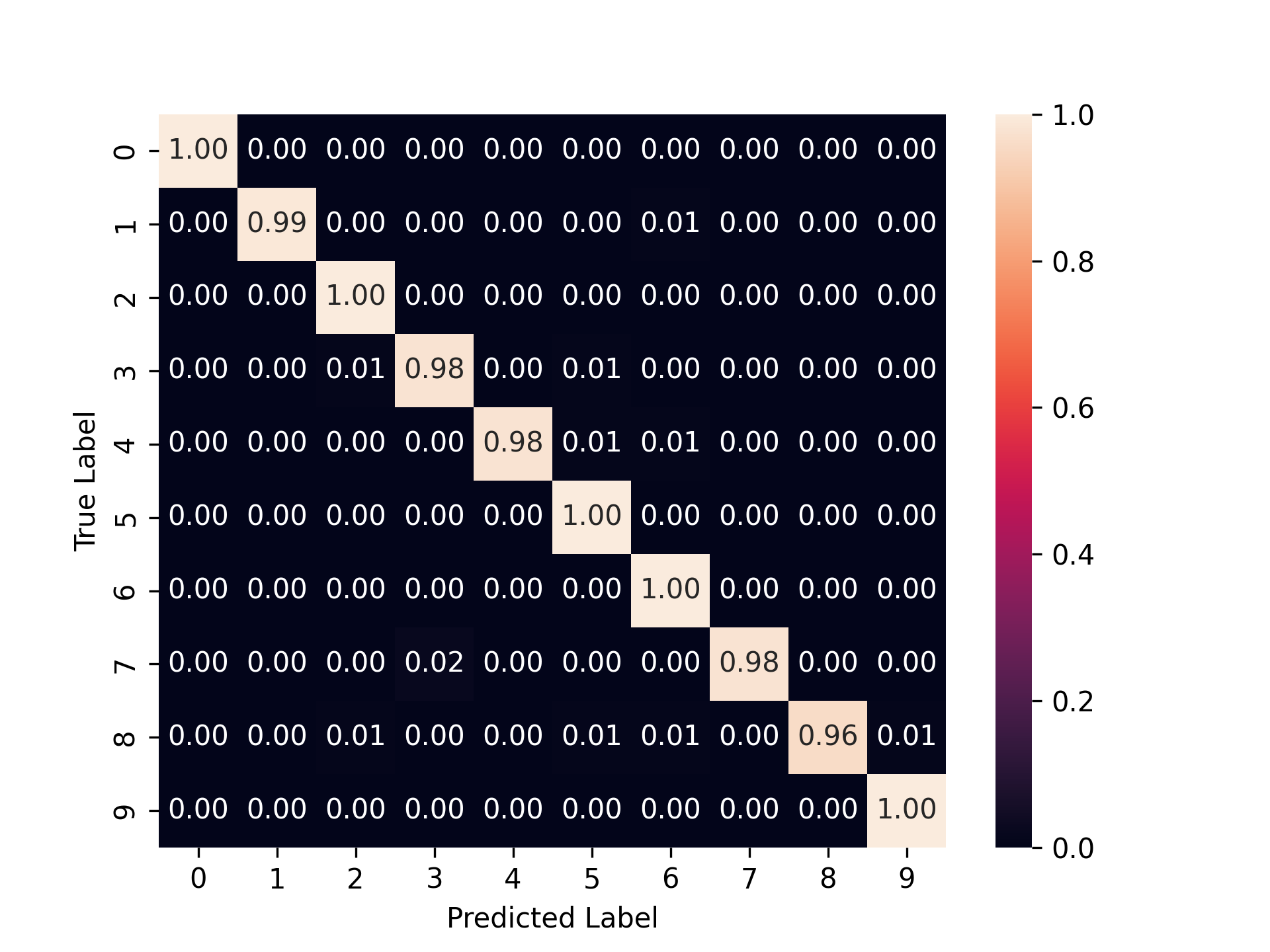} 
    \caption{Confusion matrix for diffusion model trained on the 60,000 MNIST training images. With training images corresponding to each label (row) we show the frequency with which the classifier assigned each label (column). Entries on the diagonal therefore correspond to 
    successfully created new images. Overall success rate is 99.5\%. 
    }
    \label{fig:clean}
\end{figure}

Having the trained the diffusion model, we found that 99.5\% of its  outputs were classified with the intended label.
Figure~\ref{fig:clean} gives a confusion matrix which breaks the results down by category. So, for example, for the label `7', we found that 
98\% of the outputs from the diffusion model were classified as sevens, and 2\% were classified as threes.

 \subsection{Deceptive Diffusion Model} \label{subsec:uncond}

    Our aim is now to build a \emph{deceptive diffusion model} that takes a label $i$ and 
 generates
a new image that looks like digit $i$ but is misclassified. 

\begin{figure}
    \centering
     \includegraphics[height = 0.2\textheight]{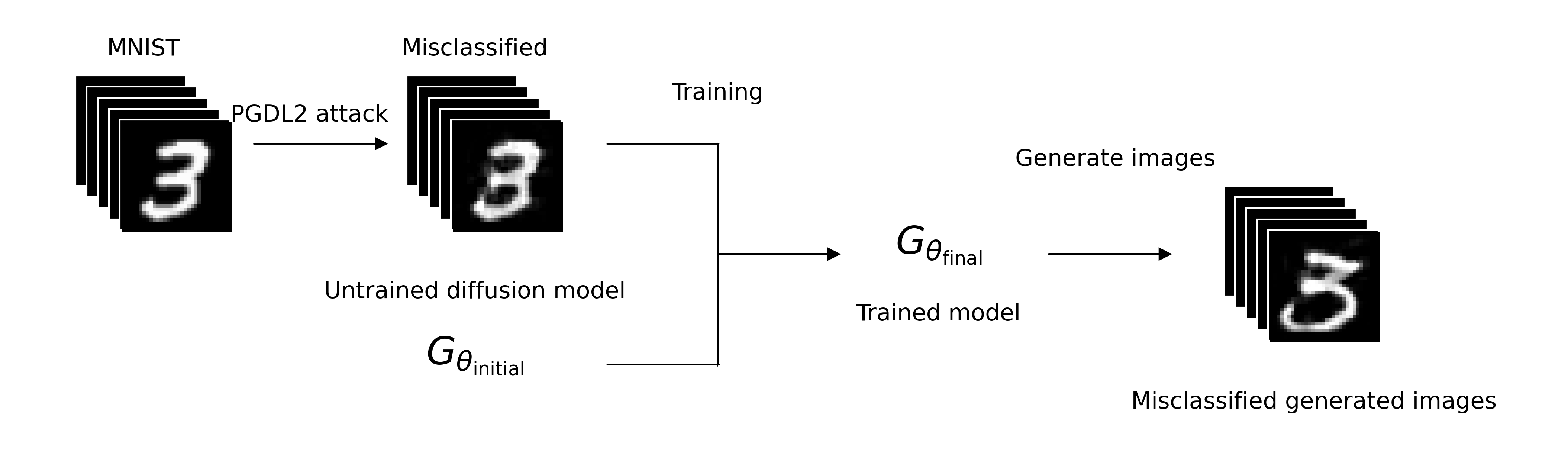} 
    \caption{Building the deceptive diffusion model.
    Images that were successfully attacked by PGDL2 are used as training data, with the original labels retained. The trained diffusion model,
    $G_{\theta_{\text{final}}}$, produces adversarial images associated with a given a label. (For the images in this diagram, the image from 
    PGDL2 is classified as an `8' and the image from the deceptive diffusion model is classified as a `5'.)
    }
    \label{fig:scheme}
\end{figure}

    Using PGDL2 for untargeted attacks on the 60,000 MNIST training images gave a success rate of
    86.5\%, thereby producing 51,918 perturbed images that are classified 
    differently to their nearby original images.
    We trained the diffusion model on these adversarial images, using the 
    original labels. Figure~\ref{fig:scheme} illustrates the
    process. Here, the image of the three on the left is from the MNIST training set, and the image in the middle arises from  
    a successful attack by PDL2 (classified as an eight). After training the diffusion model on all 51,918 adversarial images, asking for an output from the `3' category produced the result shown (classified as a five).

\begin{figure}
    \centering
  \includegraphics[height = 0.6\textheight]{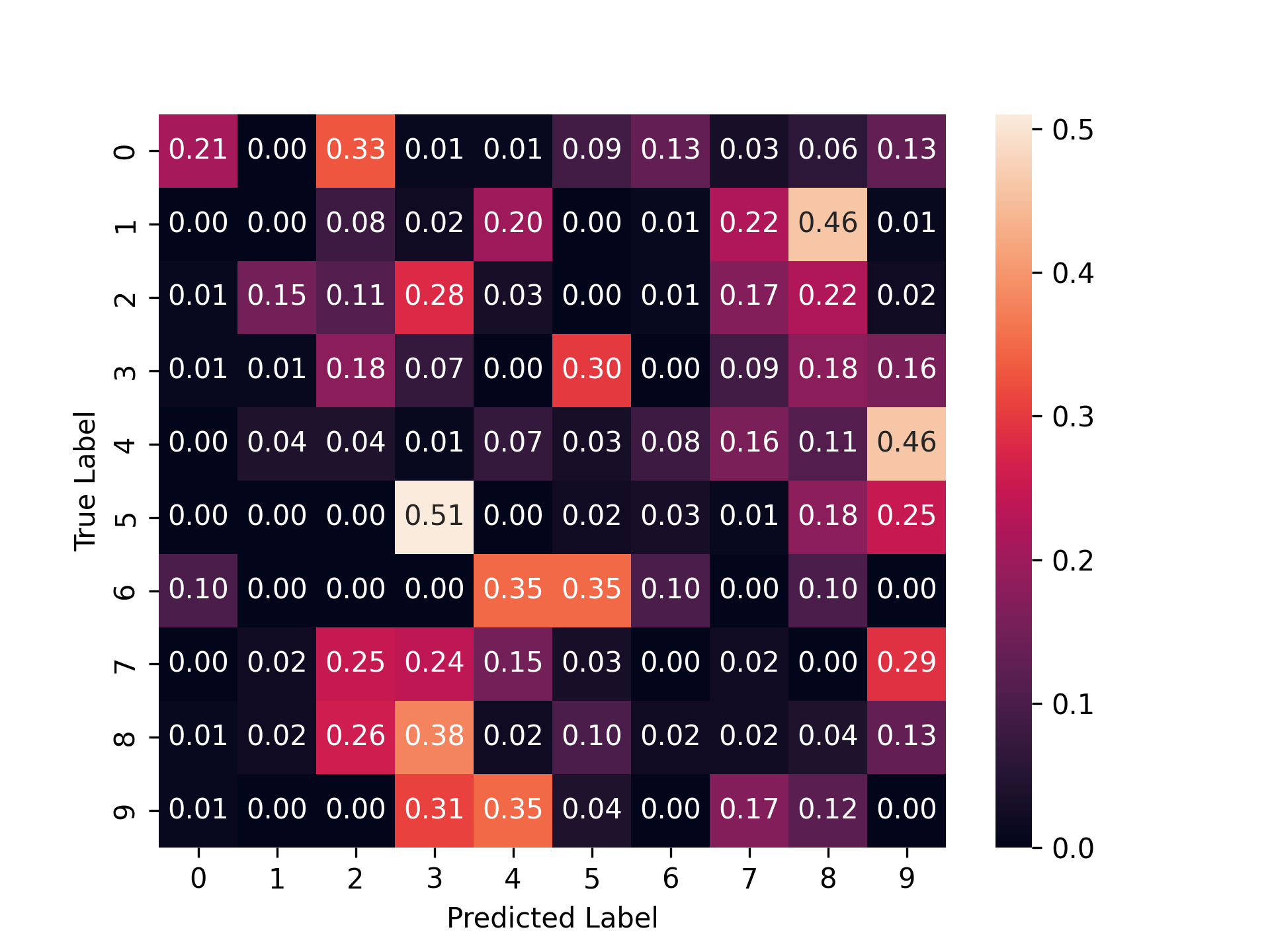} 
    \caption{Confusion matrix for the deceptive diffusion model. For a given label (row) we show the frequency with which the classifier assigned each label (column) to the output. 
    Entries on the diagonal therefore correspond to 
    unsuccessful attempts to create an adversarial image.
     Overall misclassification rate is 93.6\%.
    }
    \label{fig:uncond_diff}
\end{figure}

\begin{figure}
    \centering
     \includegraphics[height = 0.6\textheight]{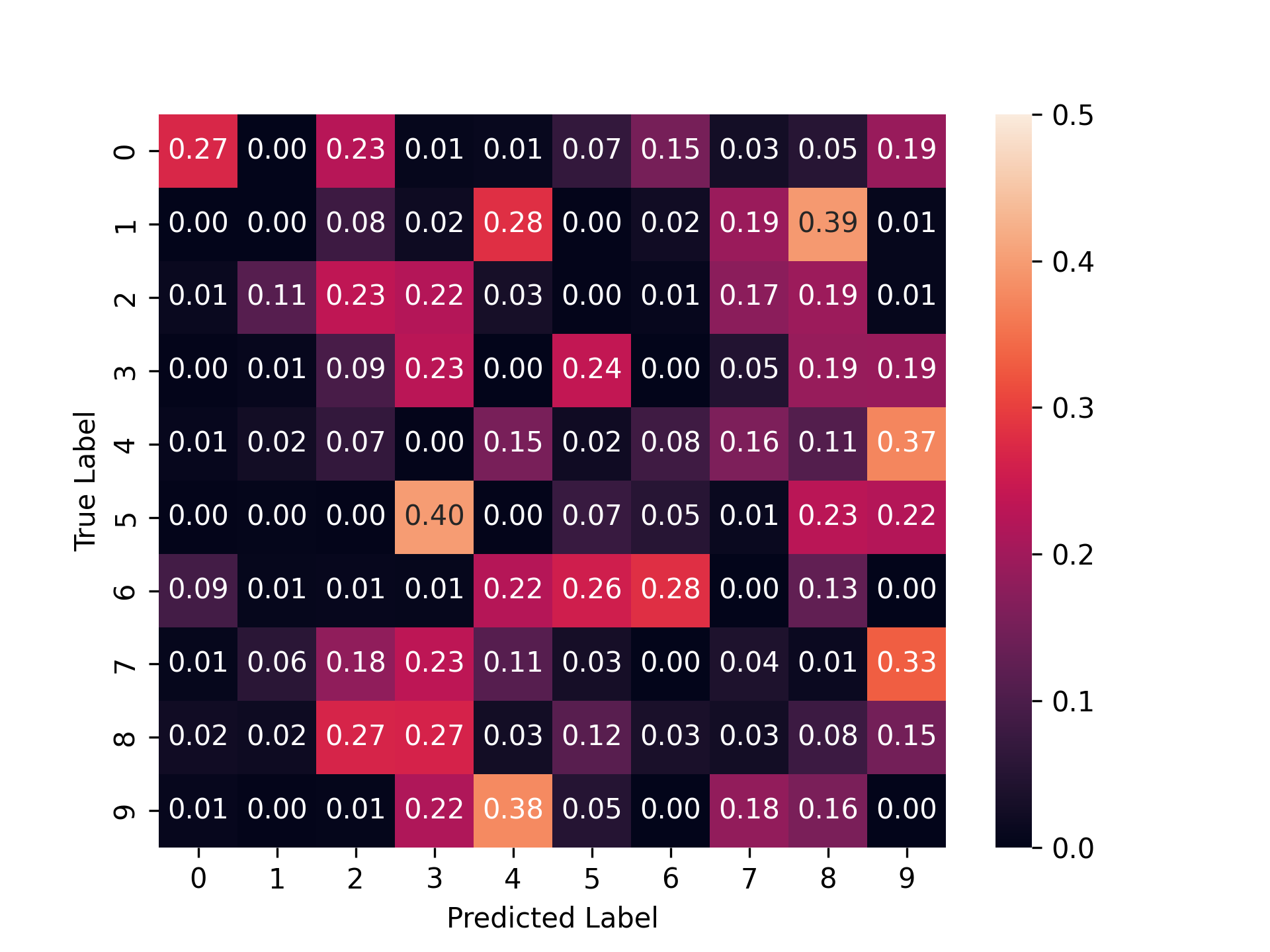} 
    \caption{
    Confusion matrix for PGDL2 attacks on the 60,000 MNIST training images. With training images corresponding to each label (row) we show the frequency with which the classifier assigned each label (column) 
    after the attack.  Entries on the diagonal therefore correspond to 
    unsuccessful attacks. Overall success rate is 86.5\%.
    }
    \label{fig:uncond_pgdl}
\end{figure}

After using the trained diffusion model to generate 100 new images from each of the ten categories and passing these through the CNN classifier,
we found that 93.6\% of the outputs were classified differently to their
requested labels. Figure~\ref{fig:uncond_diff} gives a confusion matrix showing the performance by category. For comparison,  Figure~\ref{fig:uncond_pgdl} shows a confusion matrix for the PGDL2 attacks on the 60,000 training images.

Table~\ref{tab:corr} shows the correlation between the rows of the confusion matrices in Figures~\ref{fig:uncond_diff} and \ref{fig:uncond_pgdl}. The high correlation values indicate that the two confusion matrices are similar.
We emphasize that PGDL2 was used in untargeted mode: an image from category $i$ can be perturbed so that the classifier predicts any new category $j \ne i$.
From 
Table~\ref{tab:corr} we see that although the deceptive diffusion model was not provided with the new class information $j$, it tends to produce new  
$i  \mapsto j$ misclassifications of the same type as PGDL2.

\begin{table}
    \centering
    \begin{tabular}{c||cccccccccc}
        Class & 0 &1&2&3&4&5&6&7&8&9 \\\hline
        Correlation & 0.90 & 0.97 & 0.88 & 0.79 & 0.96 & 0.98 & 0.82 & 0.96 & 0.96 & 0.96
    \end{tabular}
    \caption{Correlation of confusion matrix rows for the PGDL2 attack and the generated data.}
    \label{tab:corr}
\end{table}

\begin{figure}
    \centering
     \includegraphics[height = 0.45\textheight]{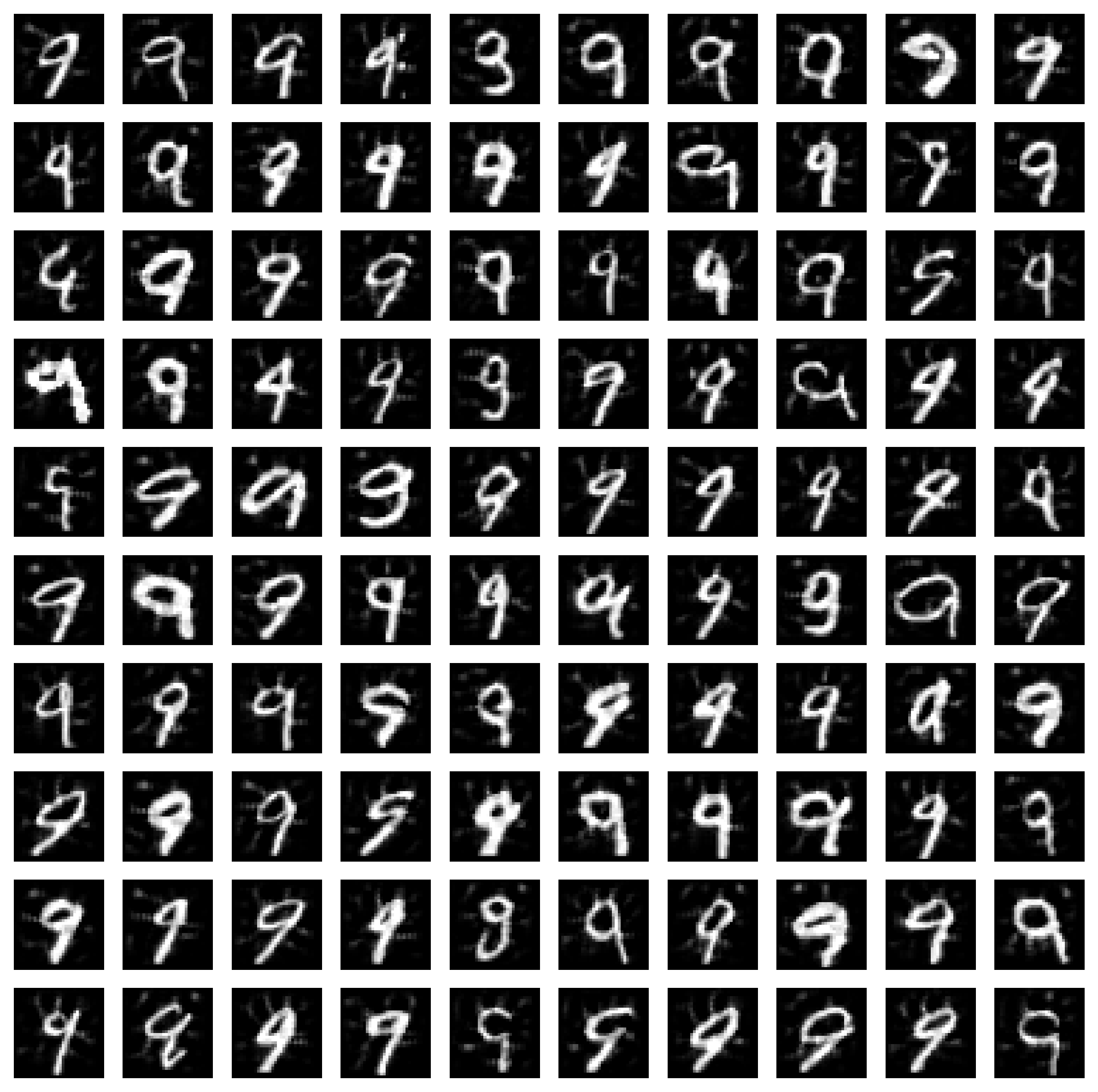}\\
\vspace{0.3in}
  \includegraphics[height = 0.45\textheight]{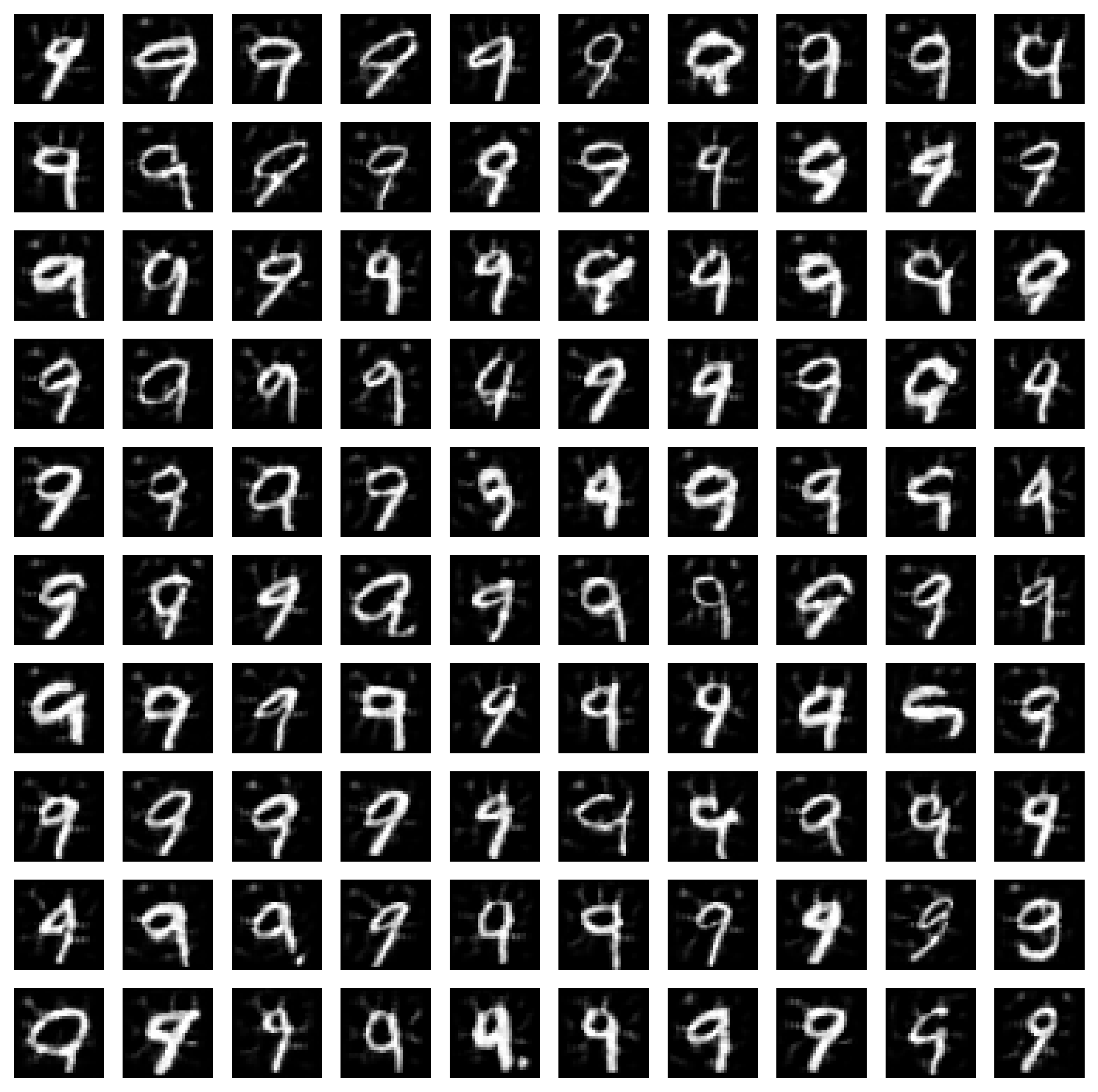} 
    \caption{Upper: 
     example of 100 images arising when the deceptive diffusion model was given the label `9'.
    Lower:
    example of 100 images arising from successful PGDL2 attacks on images that had label `9'.
    }
    \label{fig:nines}
\end{figure}

To give a feel for the outputs from the deceptive diffusion model,
Figure~\ref{fig:nines} (upper) shows 100 independent outputs corresponding to the label `9'. We note from Figure~\ref{fig:uncond_diff} that $0\%$ of such outputs are classified as nines. Hence, we see that the model is capable of producing convincing adversarial images.
For comparison, Figure~\ref{fig:nines} (lower) shows  
the results of PGDL2 on images from the `9' category. 
Similar figures for the other labels are given in  Appendix~\ref{sec:appendix}.

\subsubsection{Partial Attacks} \label{subsubsec:partial}
So far, we have looked at two options for the training data. Either all training data was attacked, or all training data was clean. Now we look at a third case: partially attacked training data. Again we choose the same MNIST images that were successfully attacked using PGDL2. Consider $p\in\{0,20,40,60,80,100\}$. For each class, we replace $p\%$ of the clean images with their successfully attacked counterpart. Now using these six datasets, we train six models. 

\begin{figure}
    \centering
    \includegraphics[width = 0.9\textwidth]{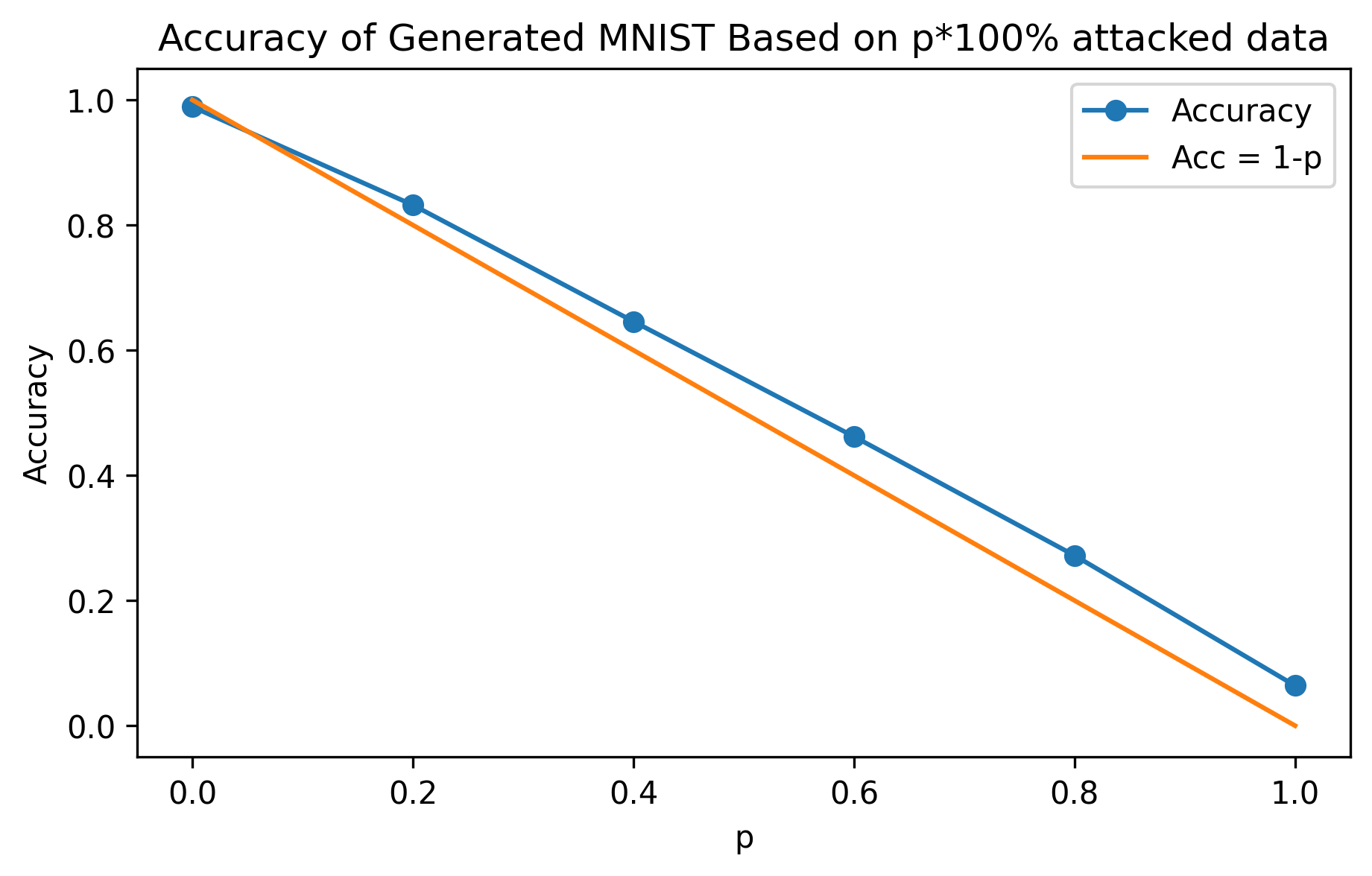} 
    \caption{
        Classification accuracy (vertical axis) for output from a deceptive diffusion model where a percentage of the training data (horizontal axis) 
        is replaced by its adversarially attacked counterpart.
        The slope representing linear proportionality is also shown.
    }
    \label{fig:partial}
\end{figure}

For each class, 100 images are generated using each of the trained models.
In Figure~\ref{fig:partial} we show the resulting accuracy of the classifier on these generated images for the models trained on varying levels of poisoned data.
We see that the classification accuracy degrades roughly in proportion 
with the amount of poisoned training data.
This result is intuitively reasonable, under the assumption that 
all training images carry equal weight when the diffusion model 
is created. Confusion matrices for the partially trained models can be found in Appendix~\ref{sec:appendix_confusion}.

\subsubsection{Fr{\'e}chet Inception Distance} \label{subsubsec:incept}
A widely used measure for generated image quality is the Fr{\'e}chet Inception Distance (FID)\cite{Heu17}, where lower is better. It compares a generated dataset to a ground truth dataset. First, a classifier is used to extract features. Then the Fr{\'e}chet distance between these feature sets is computed. Typically the Inception v3 classifier \cite{szegedy2016rethinking} without its last layer is used. To take into account that the generator is conditioned on the class, we use the Class-Aware Fr{\'e}chet Distance (CAFD), which computes the FID for every class and takes the average \cite{liu2018improved}. 

Since our dataset is of low resolution, instead of Inception v3 we use the classifier that we trained earlier, with its last layer removed. This way the output is in $\mathbb{R}^{128}$. 

\begin{figure}
    \centering
    \includegraphics[width = 0.8\textwidth]{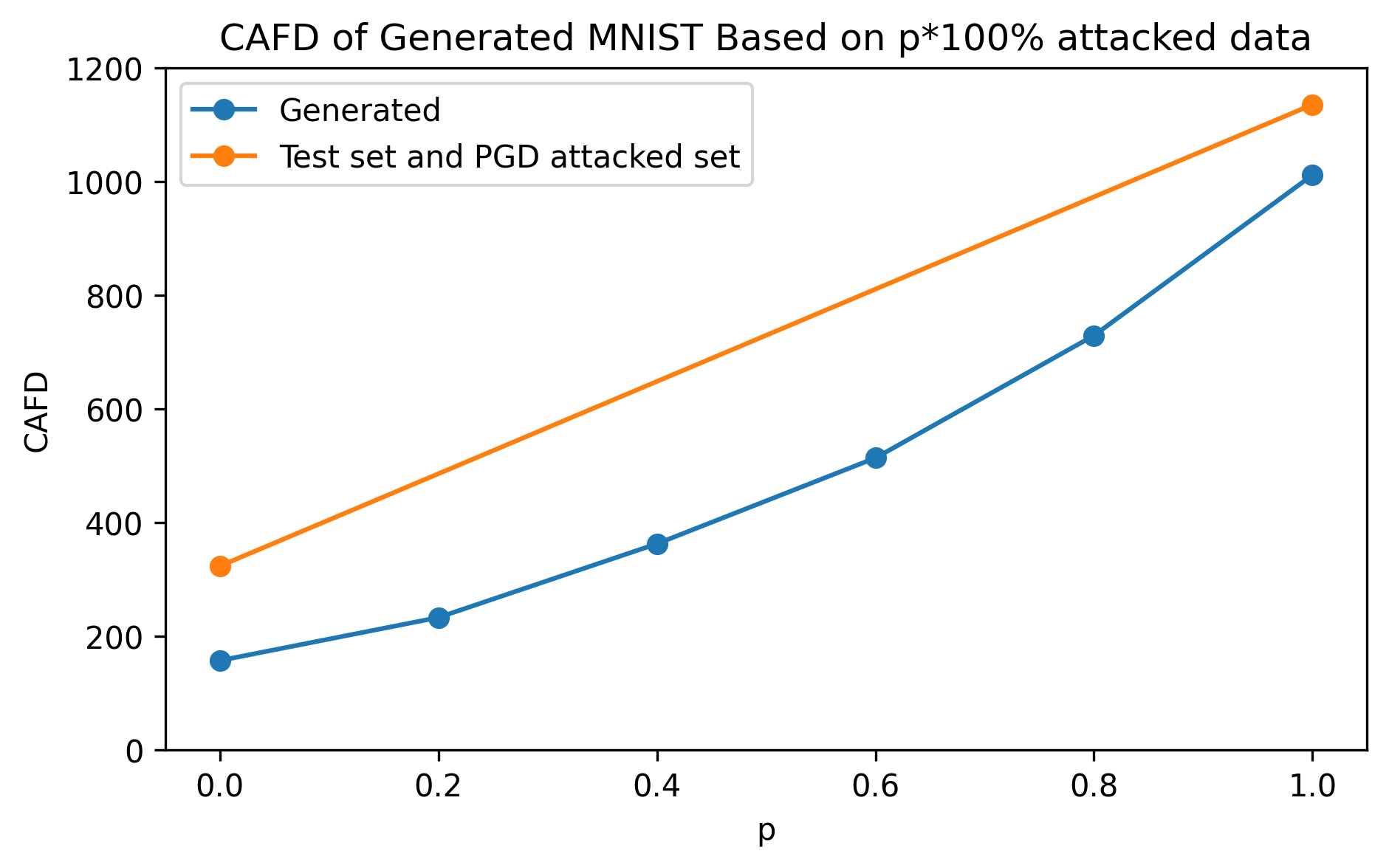}
    \caption{Class-aware Fr{\'e}chet Distance for a deceptive diffusion model where a percentage of the training data (horizontal axis) is replaced by its adversarially attacked counterpart. The ground truth dataset is MNIST. The straight line joins the CAFD for the test set at $p=0$ and the PGDL2 attacked training set at $p=1$. These two sets contain the same number of samples as the generated sets.}
    \label{fig:cafd}
\end{figure}

In Figure~\ref{fig:cafd}, the CAFD is shown for the diffusion models trained with partially poisoned data. These values are compared with the CAFD for the test set  and the PGDL2 attacked training set. These are displayed at $p=0$ and $p=1$ respectively, because they represent samples from the ground truths for the clean and attacked case respectively. 
To avoid bias, 
these two sets are limited to contain the same number of samples as the generated sets, \cite{chong2020effectively}.

The results in Figure~\ref{fig:cafd} show that the CAFD increases 
monotonically as the level of poisoning increases. This seems reasonable, because,
as shown in 
Figure~\ref{fig:partial},
higher levels of poisoning lead to higher levels of misclassification. The CAFD relies on the feature extraction of an MNIST classifier. Since the attacks target the classifier, it makes sense that the extracted features are different. 
The key observation here is that the fully adversarial model ($p=1$) corresponds to a CAFD that is similar to that of the PGDL2 attacked data set,
indicating that deception diffusion can mimic adversarially attacked 
data successfully according to this metric.

\section{Conclusions}\label{sec:conc}
 
 A traditional adversarial attack algorithm aims to perturb an existing image across a decision boundary. 
 Instead, by training a
 generative diffusion model on adversarial data, we are able to 
 create synthetic images that automatically 
 lie on the wrong side of a decision boundary.
  This observation, which we believe to have been made for the first time in this work, reveals a new type of vulnerability for generative AI:
  if a diffusion model is inadvertently trained on fully or partially
  poisoned data then a tool may be produced that generates unlimited amounts of classifier-fooling examples.

 In common with the AdvDiffuser algorithm in \cite{Chen_2023_ICCV}, 
 when deliberately trained on adversarial data, 
 a \emph{deceptive diffusion} model has the potential to  
 \begin{itemize}
     \item  create effective adversarial images at scale, independently of the amount of training and test data available,
       \item  create examples of misclassification that are difficult to obtain with a traditional adversarial attack; for example, in a healthcare setting when certain classes are underrepresented in the data \cite{Kte24}.
 \end{itemize}
This technique has applications for defence as well as attack, since it provides valuable new sources of data for adversarial training algorithms that aim to improve robustness. 

 There are many directions in which the deceptive diffusion idea could be pursued; notably, testing on other types of labeled image data, 
 generating adversarial images that are successful across a range of 
 independent classifiers, and finding computable signatures with which to identify this new type of threat.

\section*{Data Statement}
Code for these experiments will be made available upon publication.

\bibliographystyle{siam}
\bibliography{diff_refs}

\appendix
\section{Further Output Examples}\label{sec:appendix}
\begin{figure}
    \centering
     \includegraphics[height = 0.45\textheight]{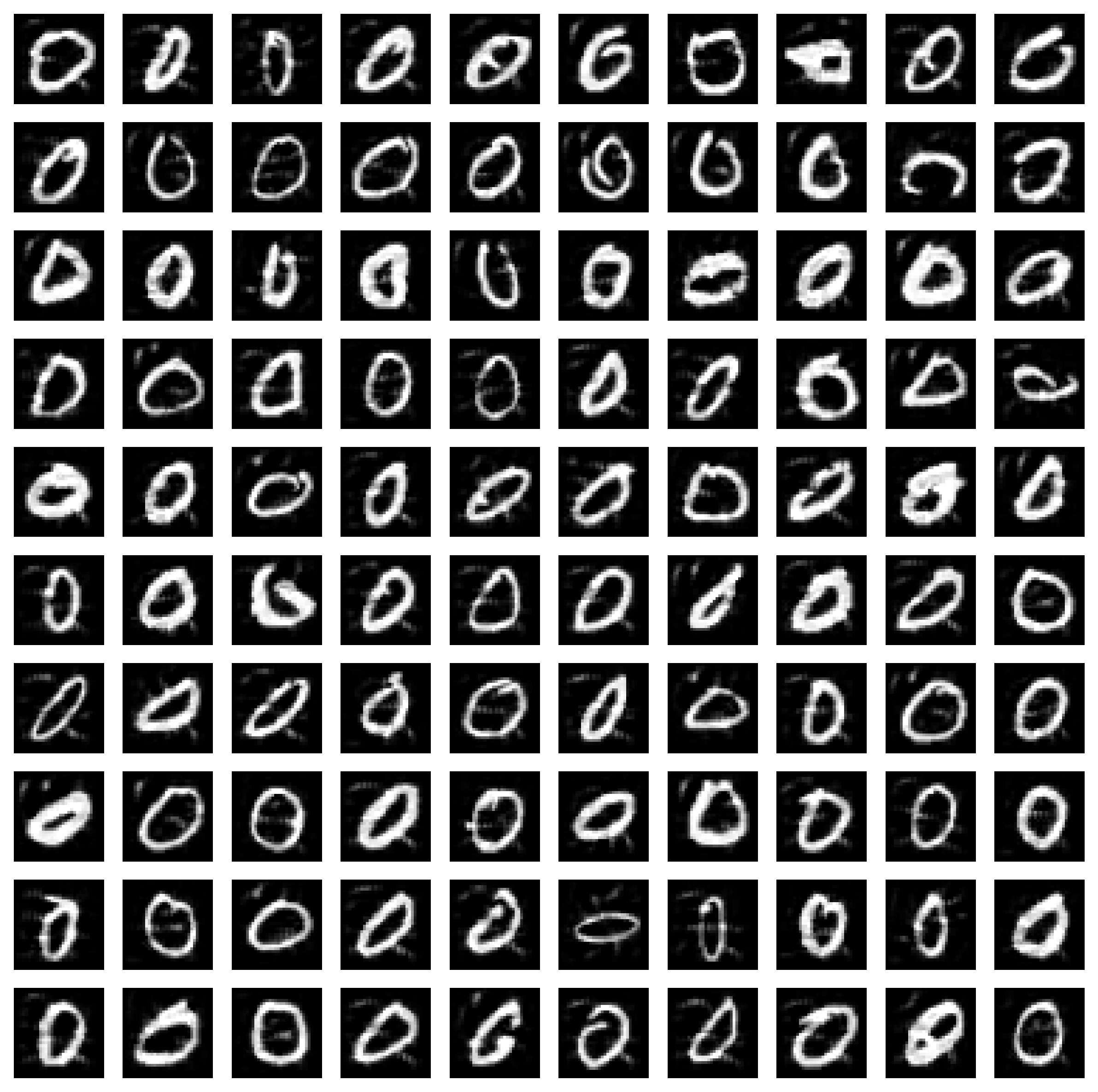}\\
\vspace{0.3in}
  \includegraphics[height = 0.45\textheight]{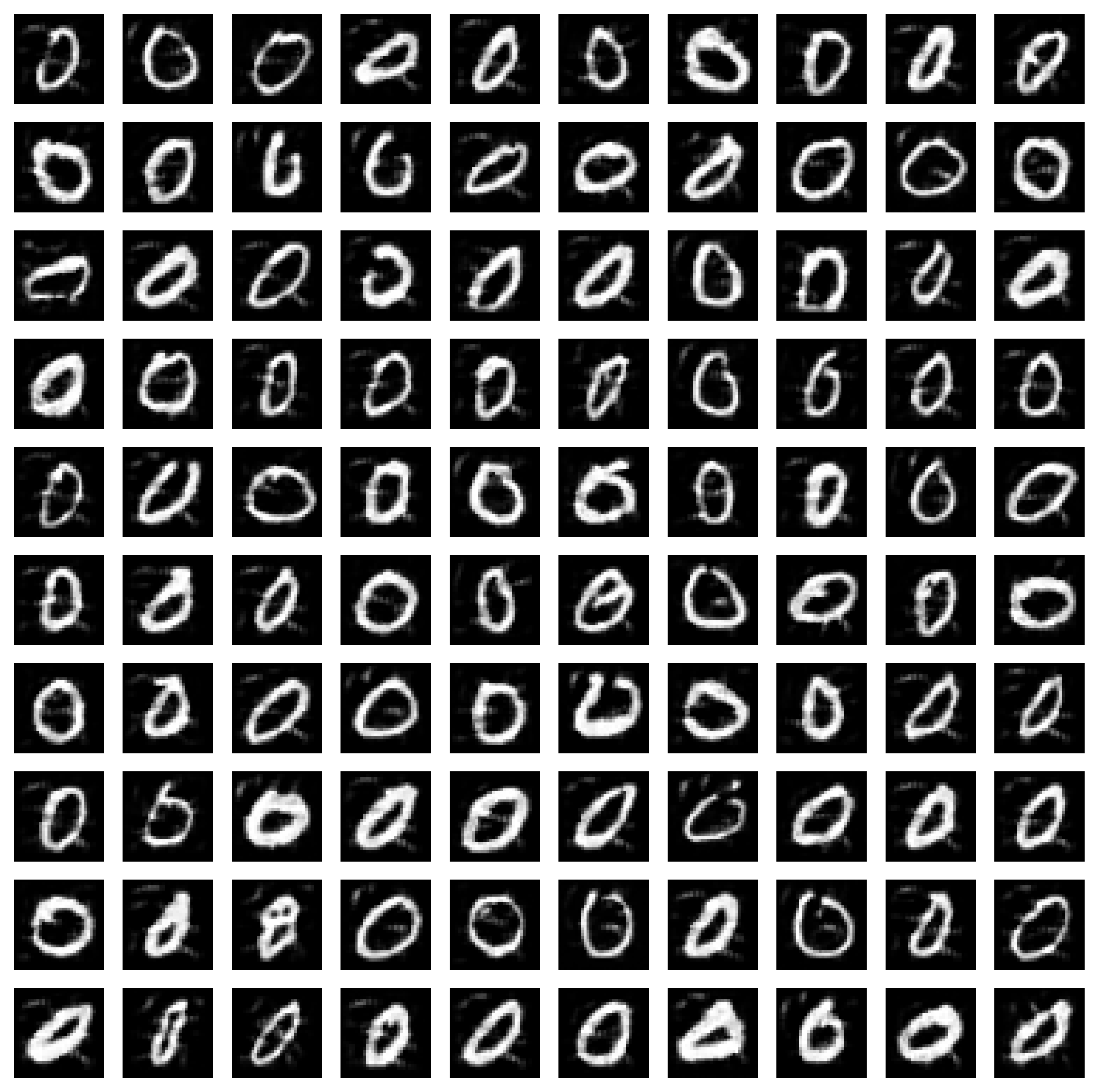} 
    \caption{Upper: 
     example of 100 images arising when the deceptive diffusion model was given the label `0'.
    Lower:
    example of 100 images arising from successful PGDL2 attacks on images that had label `0'.
    }
    \label{fig:grid0}
\end{figure}

\begin{figure}
    \centering
     \includegraphics[height = 0.45\textheight]{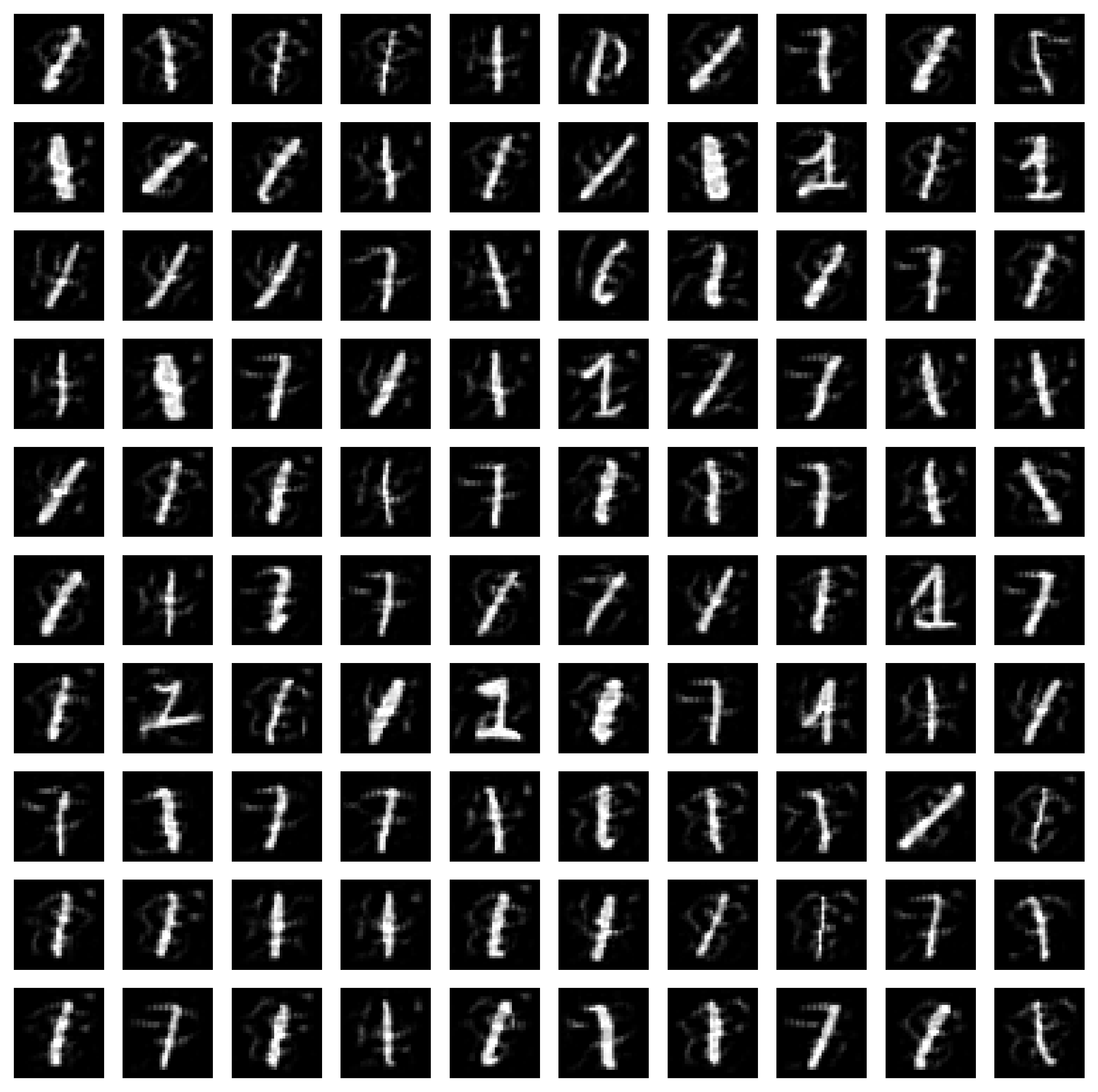}\\
\vspace{0.3in}
  \includegraphics[height = 0.45\textheight]{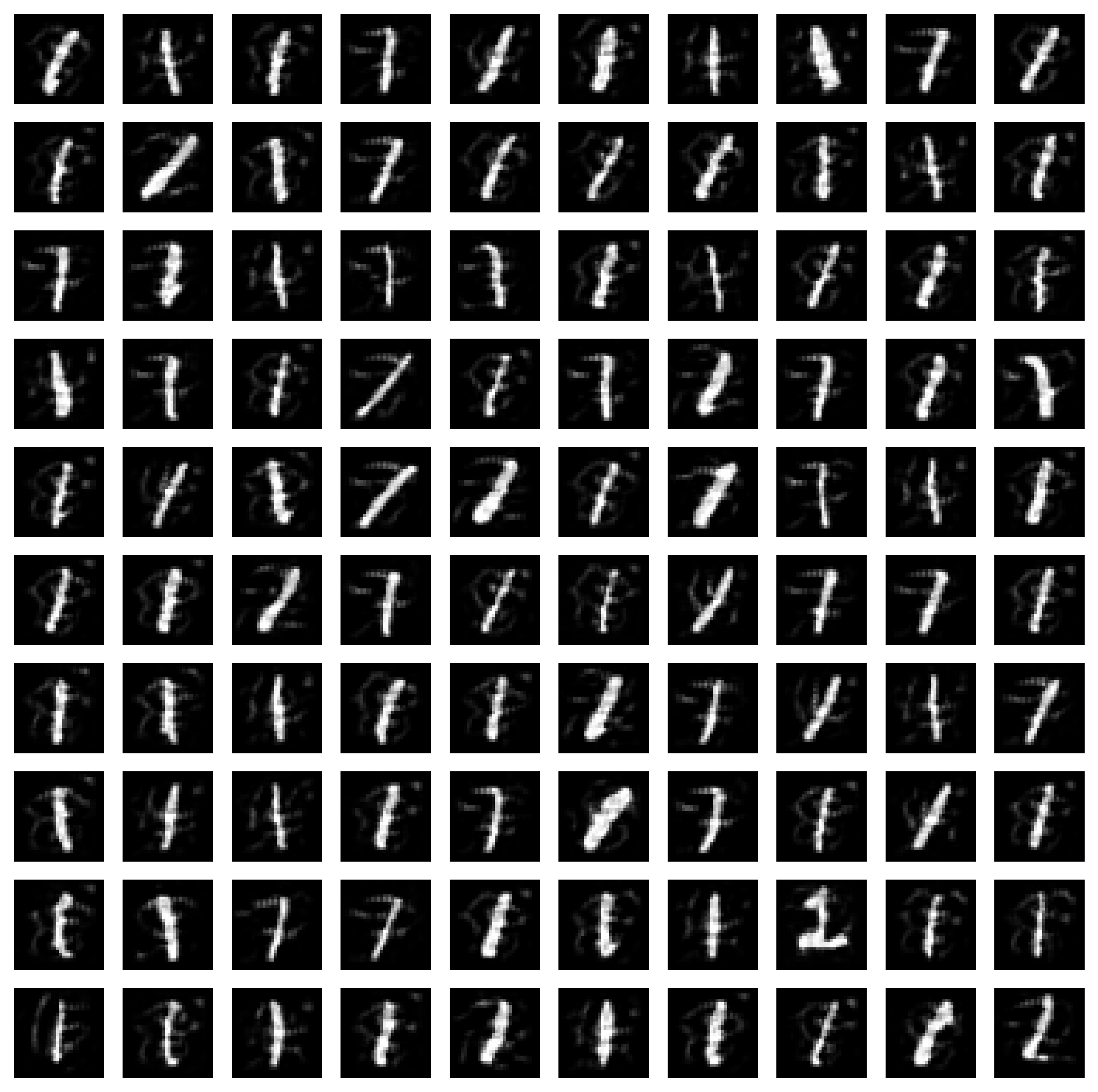} 
    \caption{Upper: 
     example of 100 images arising when the deceptive diffusion model was given the label `1'.
    Lower:
    example of 100 images arising from successful PGDL2 attacks on images that had label `1'.
    }
    \label{fig:grid1}
\end{figure}

\begin{figure}
    \centering
     \includegraphics[height = 0.45\textheight]{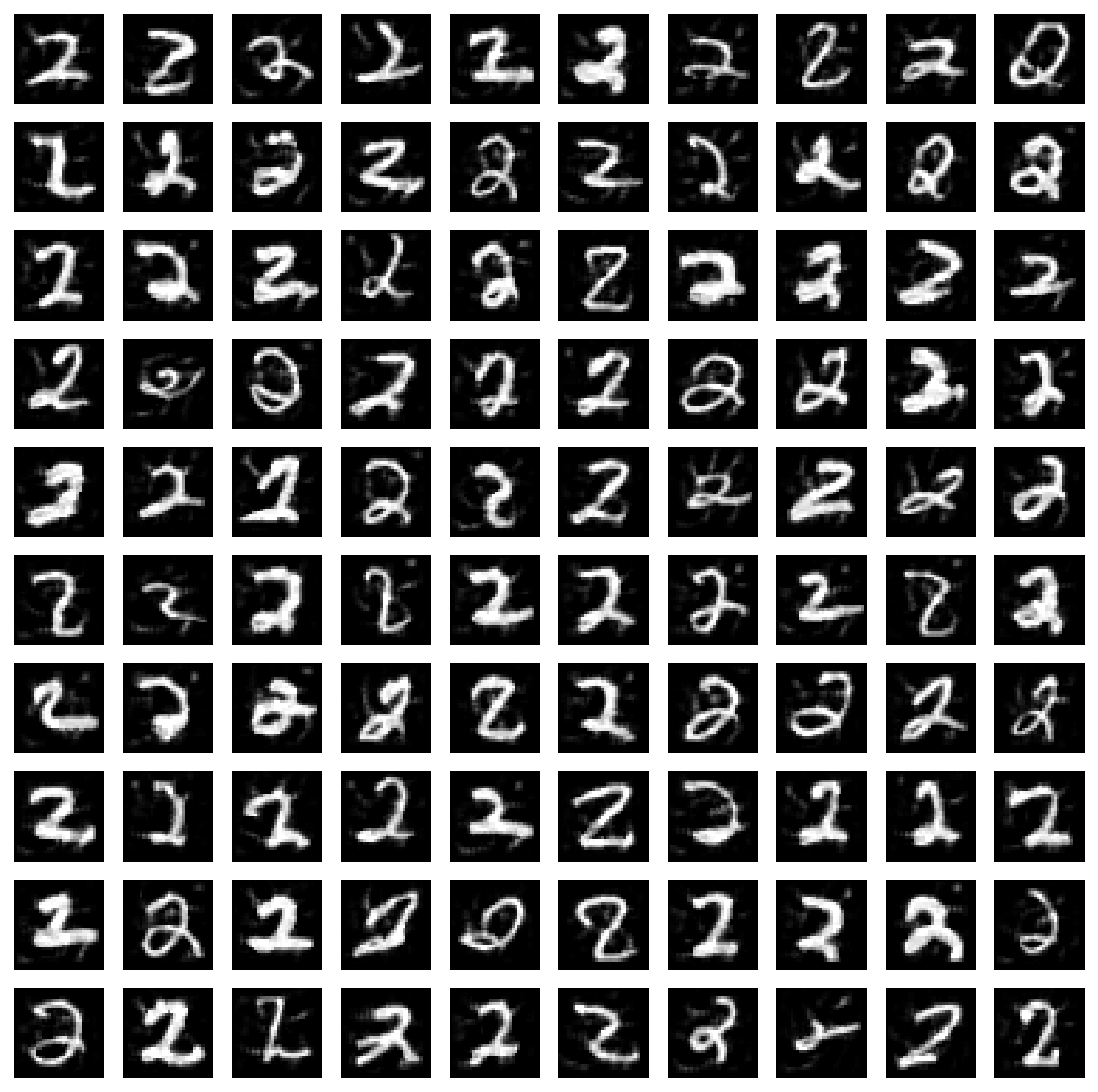}\\
\vspace{0.3in}
  \includegraphics[height = 0.45\textheight]{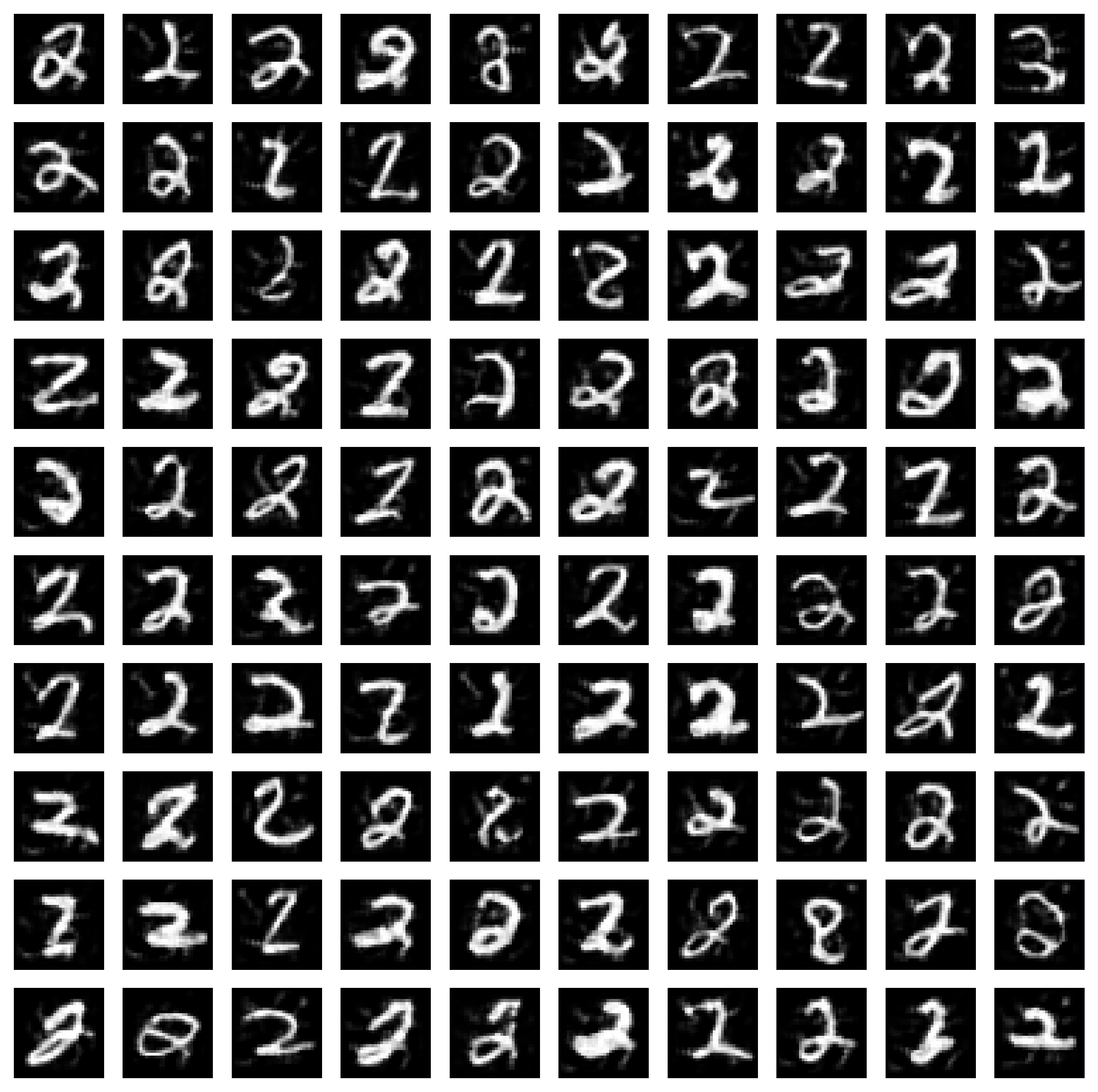} 
    \caption{Upper: 
     example of 100 images arising when the deceptive diffusion model was given the label `2'.
    Lower:
    example of 100 images arising from successful PGDL2 attacks on images that had label `2'.
    }
    \label{fig:grid2}
\end{figure}

\begin{figure}
    \centering
     \includegraphics[height = 0.45\textheight]{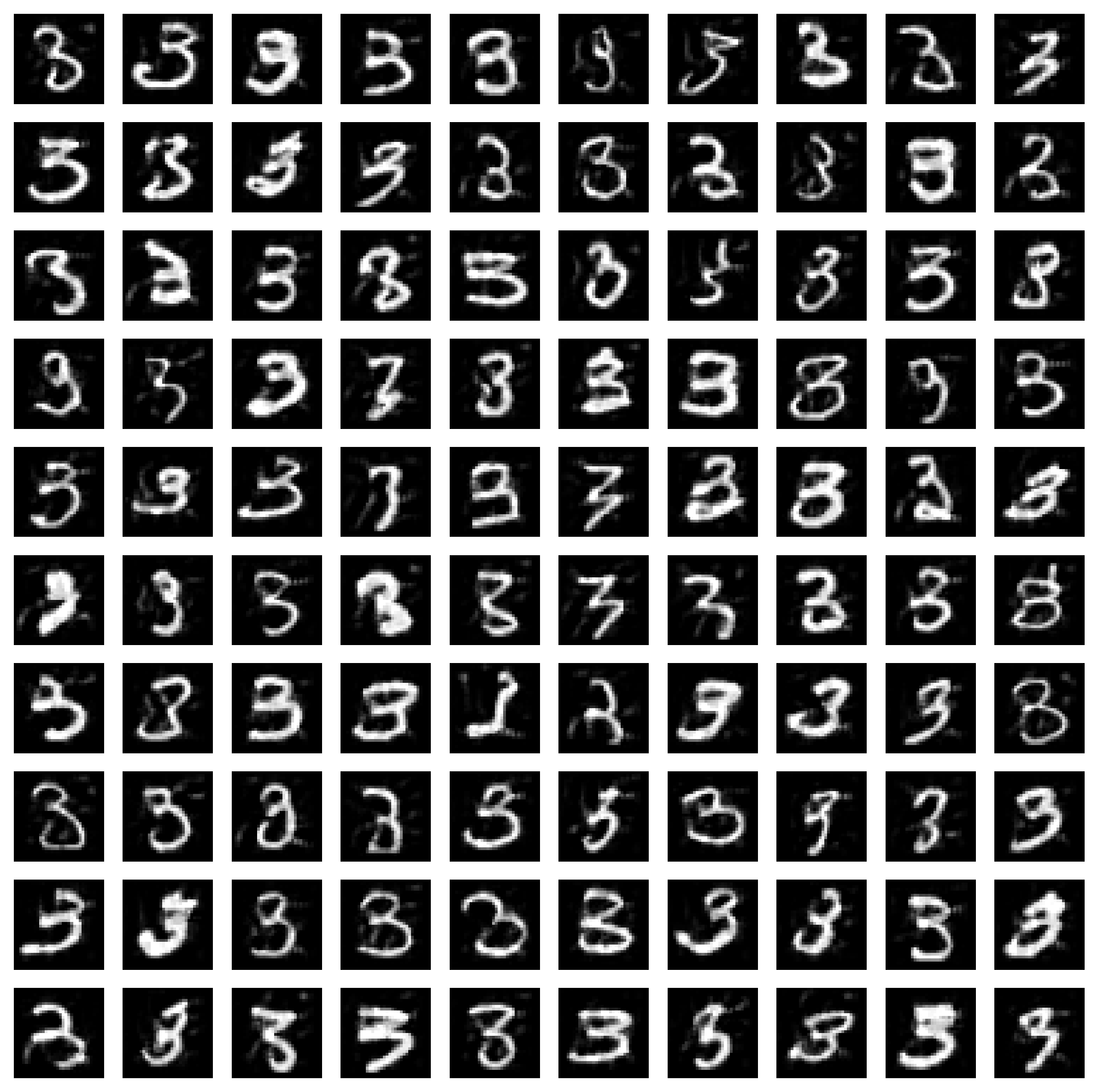}\\
\vspace{0.3in}
  \includegraphics[height = 0.45\textheight]{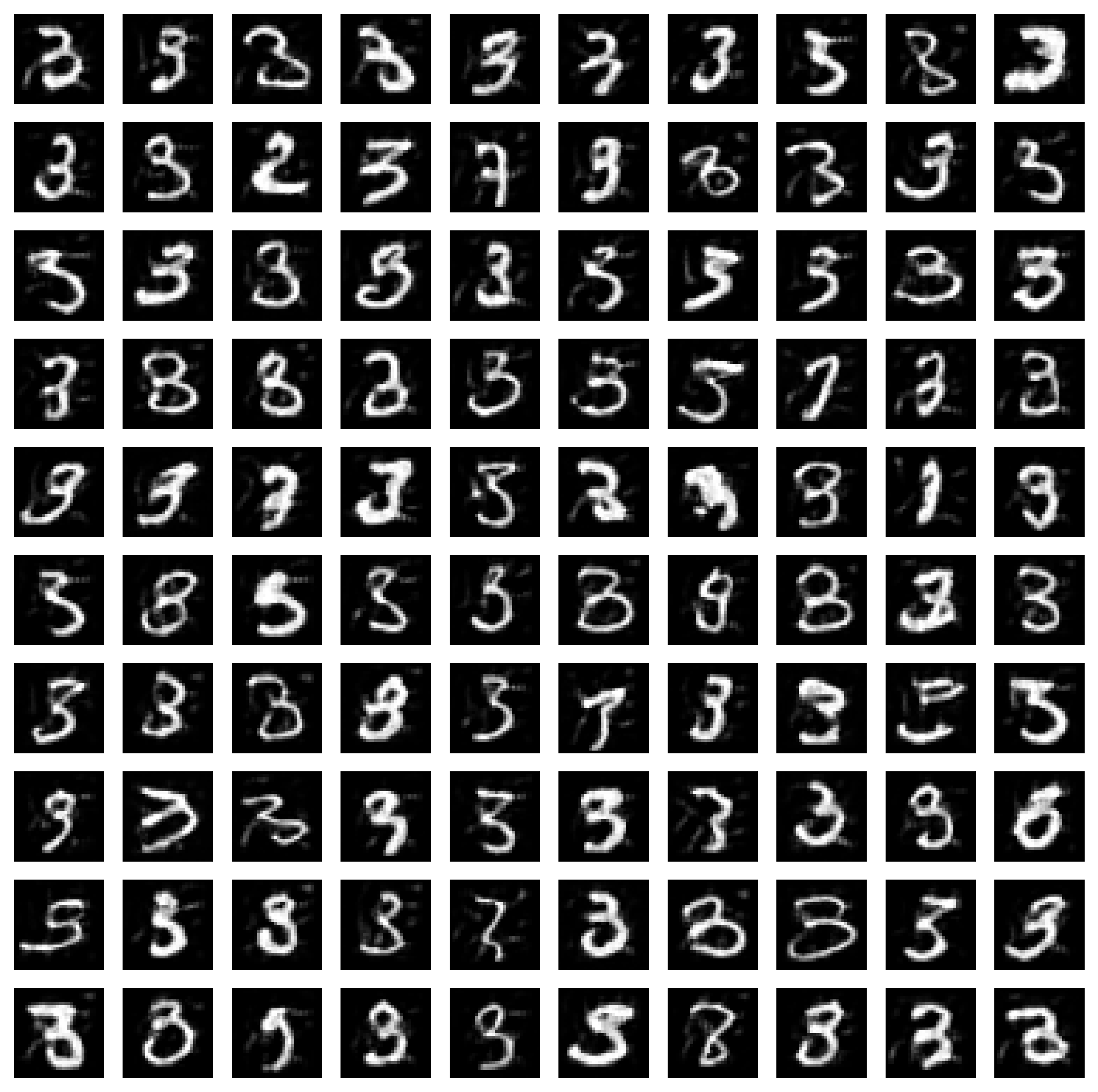} 
    \caption{Upper: 
     example of 100 images arising when the deceptive diffusion model was given the label `3'.
    Lower:
    example of 100 images arising from successful PGDL2 attacks on images that had label `3'.
    }
    \label{fig:grid3}
\end{figure}

\begin{figure}
    \centering
     \includegraphics[height = 0.45\textheight]{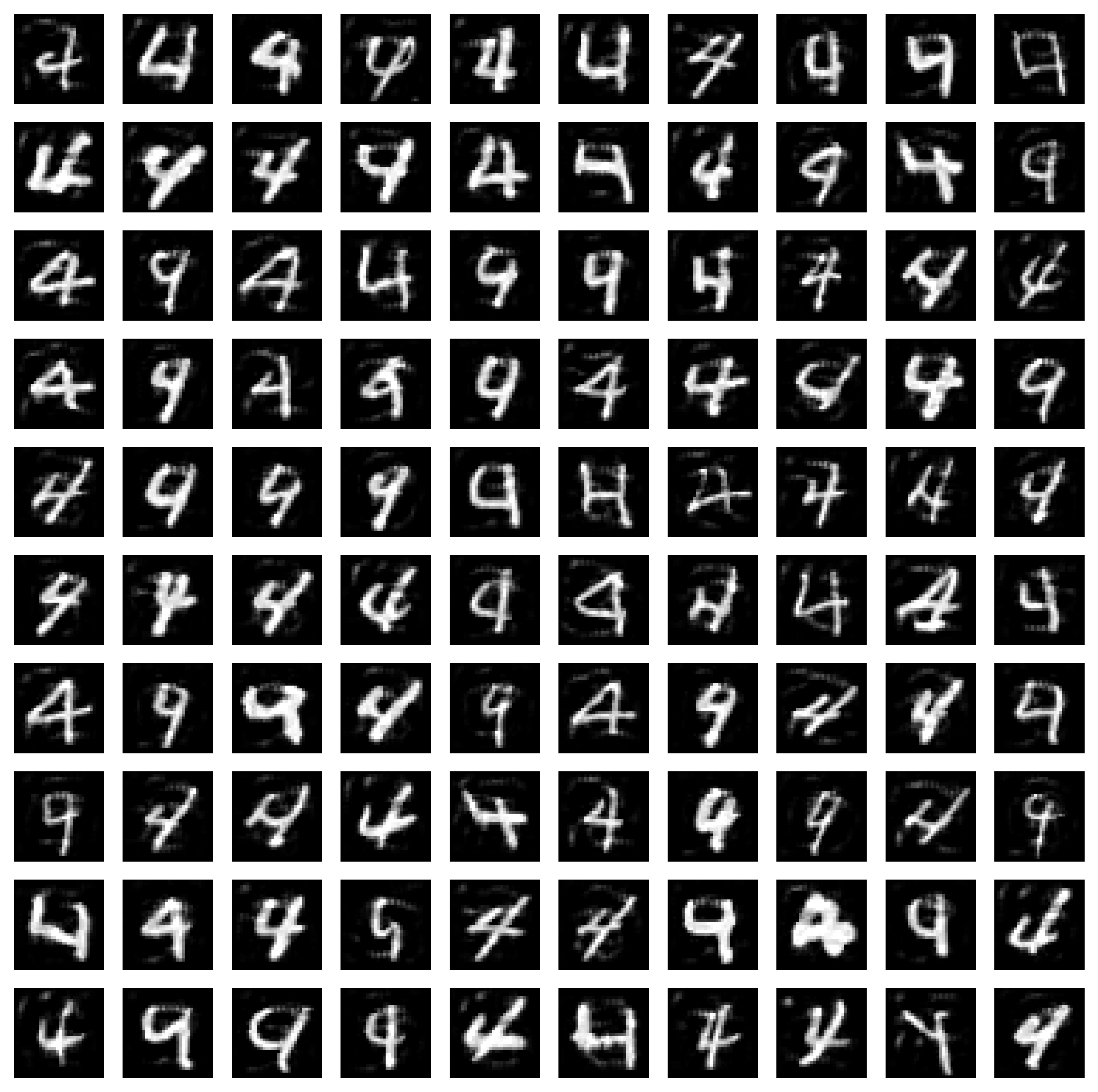}\\
\vspace{0.3in}
  \includegraphics[height = 0.45\textheight]{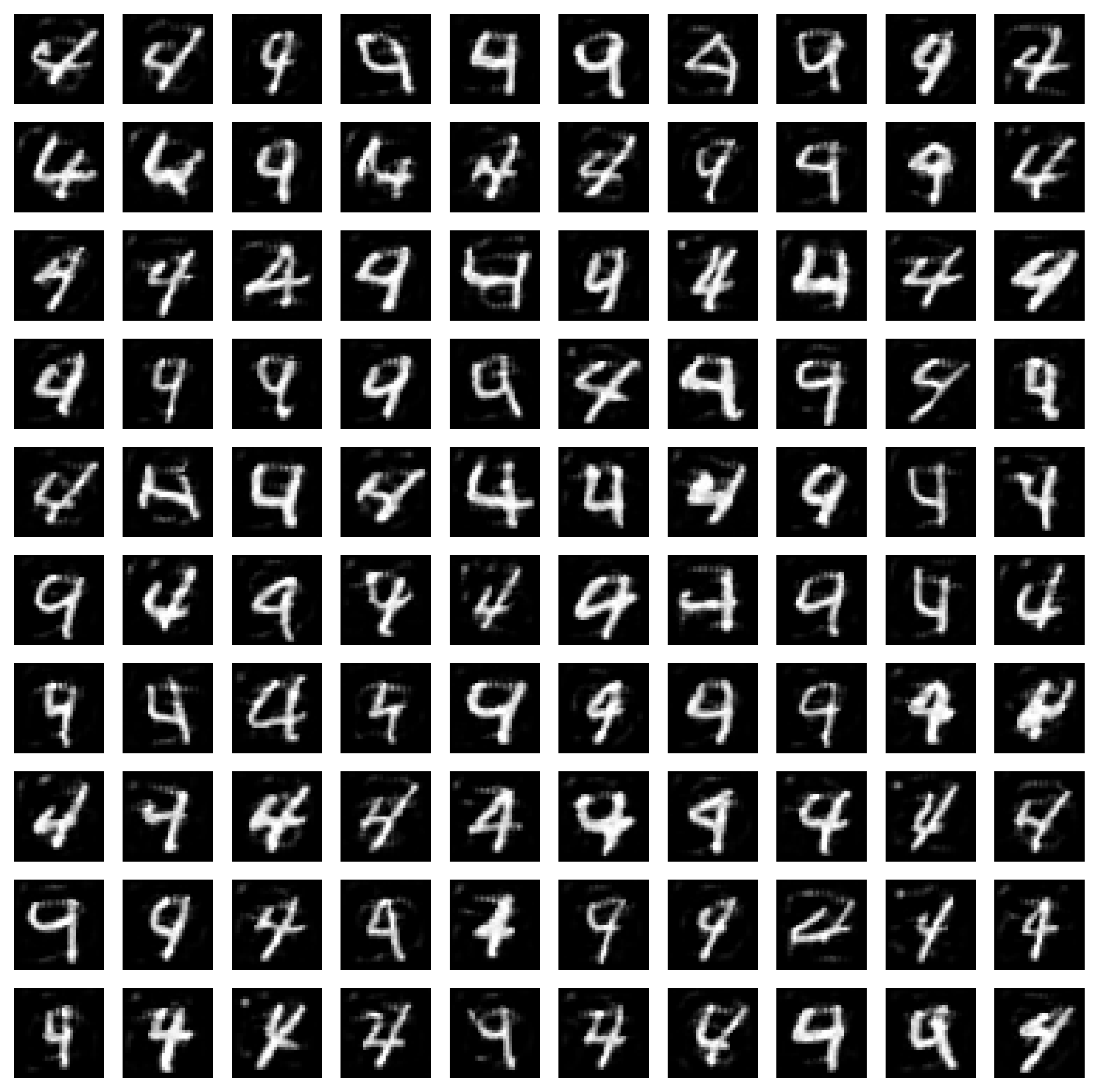} 
    \caption{Upper: 
     example of 100 images arising when the deceptive diffusion model was given the label `4'.
    Lower:
    example of 100 images arising from successful PGDL2 attacks on images that had label `4'.
    }
    \label{fig:grid4}
\end{figure}

\begin{figure}
    \centering
     \includegraphics[height = 0.45\textheight]{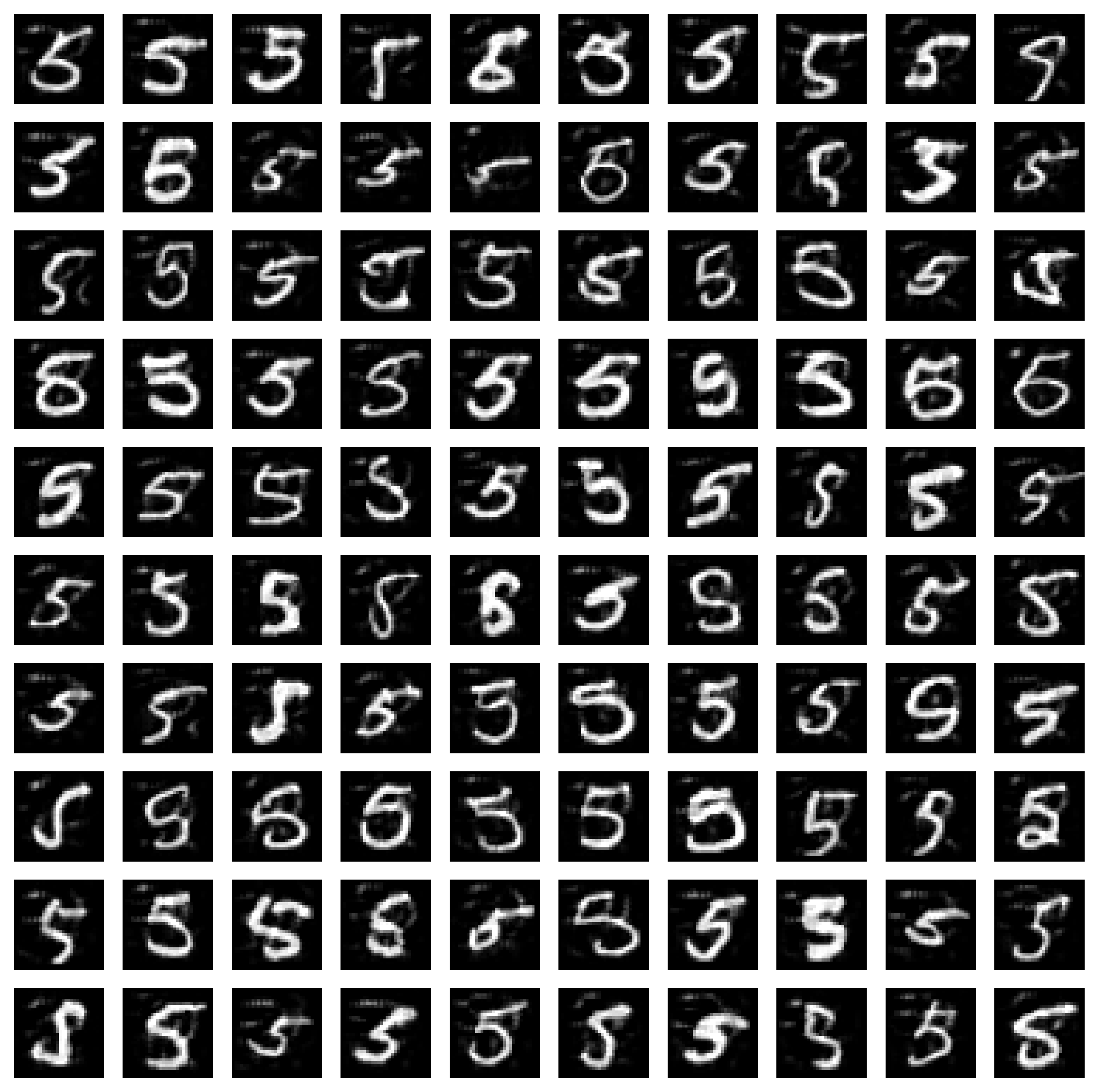}\\
\vspace{0.3in}
  \includegraphics[height = 0.45\textheight]{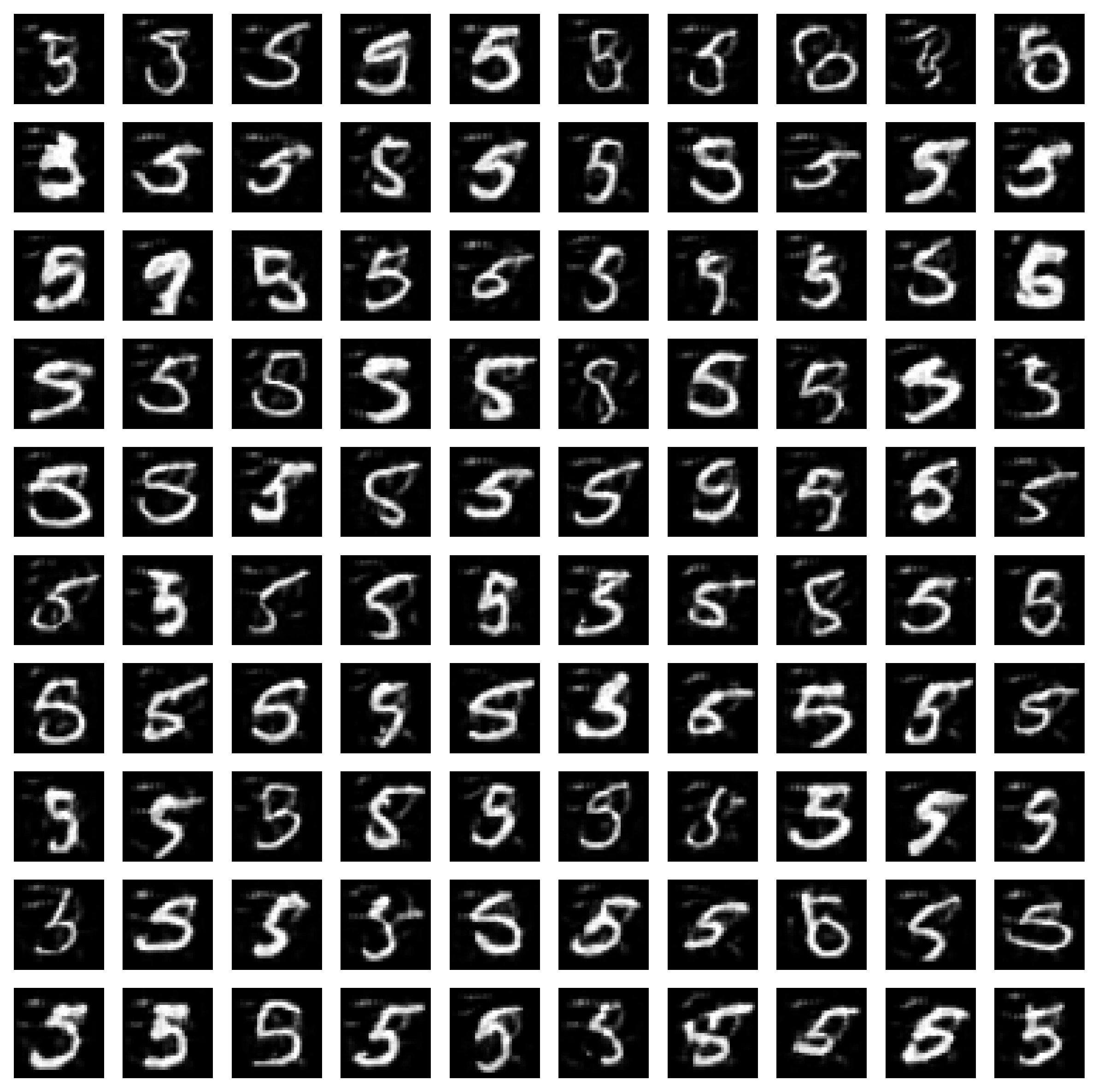} 
    \caption{Upper: 
     example of 100 images arising when the deceptive diffusion model was given the label `5'.
    Lower:
    example of 100 images arising from successful PGDL2 attacks on images that had label `5'.
    }
    \label{fig:grid5}
\end{figure}

\begin{figure}
    \centering
     \includegraphics[height = 0.45\textheight]{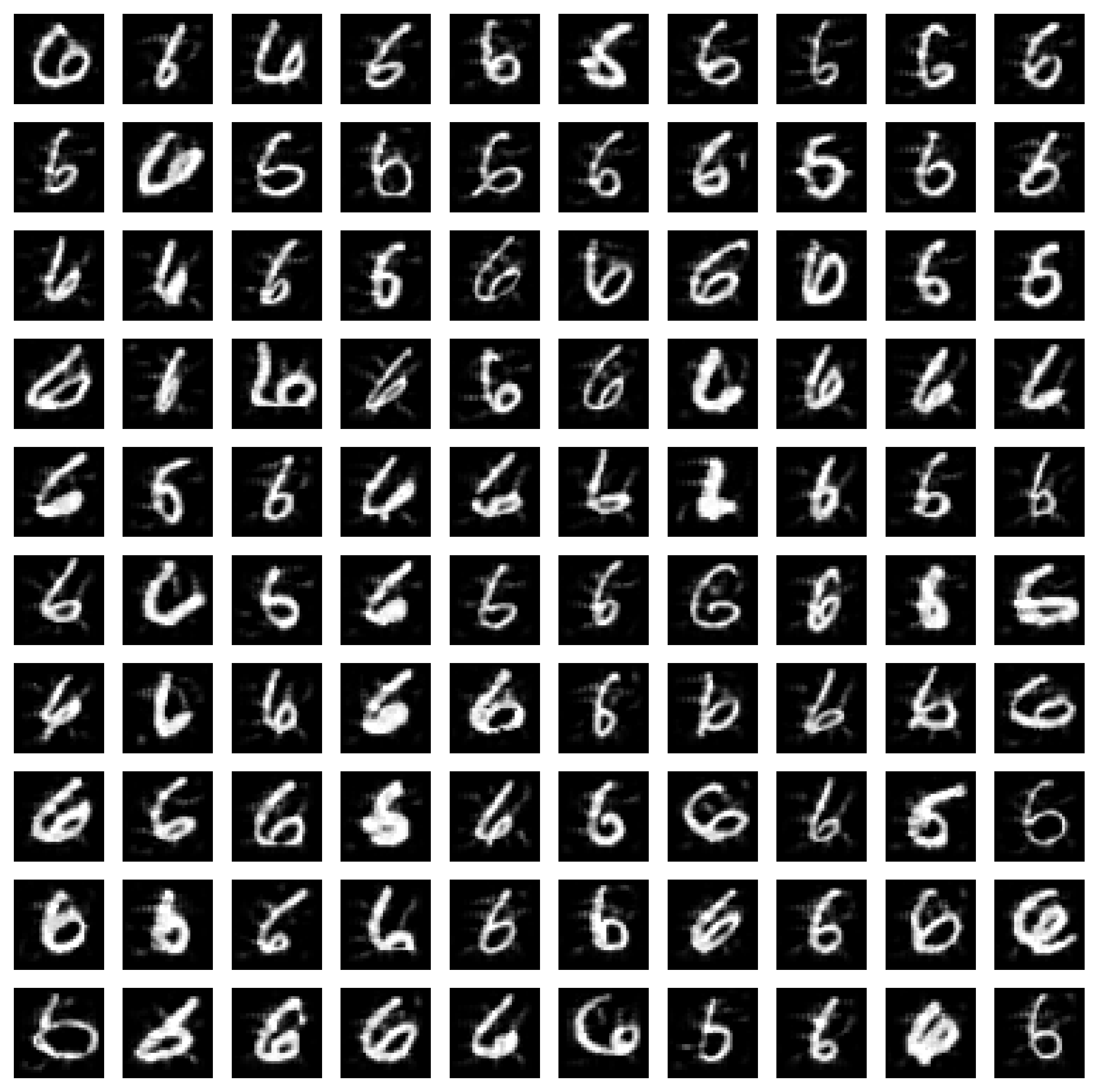}\\
\vspace{0.3in}
  \includegraphics[height = 0.45\textheight]{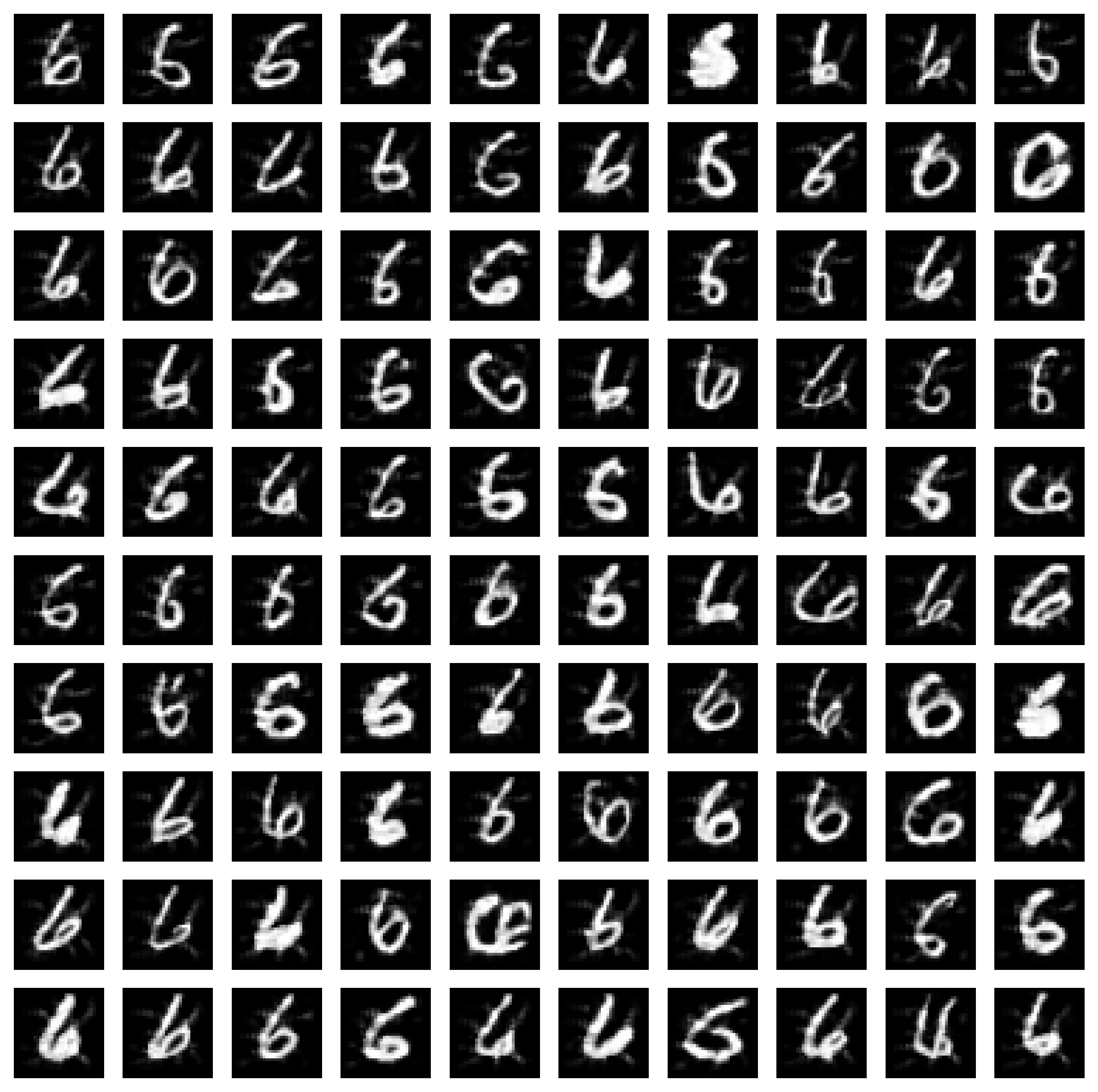} 
    \caption{Upper: 
     example of 100 images arising when the deceptive diffusion model was given the label `6'.
    Lower:
    example of 100 images arising from successful PGDL2 attacks on images that had label `6'.
    }
    \label{fig:grid6}
\end{figure}

\begin{figure}
    \centering
     \includegraphics[height = 0.45\textheight]{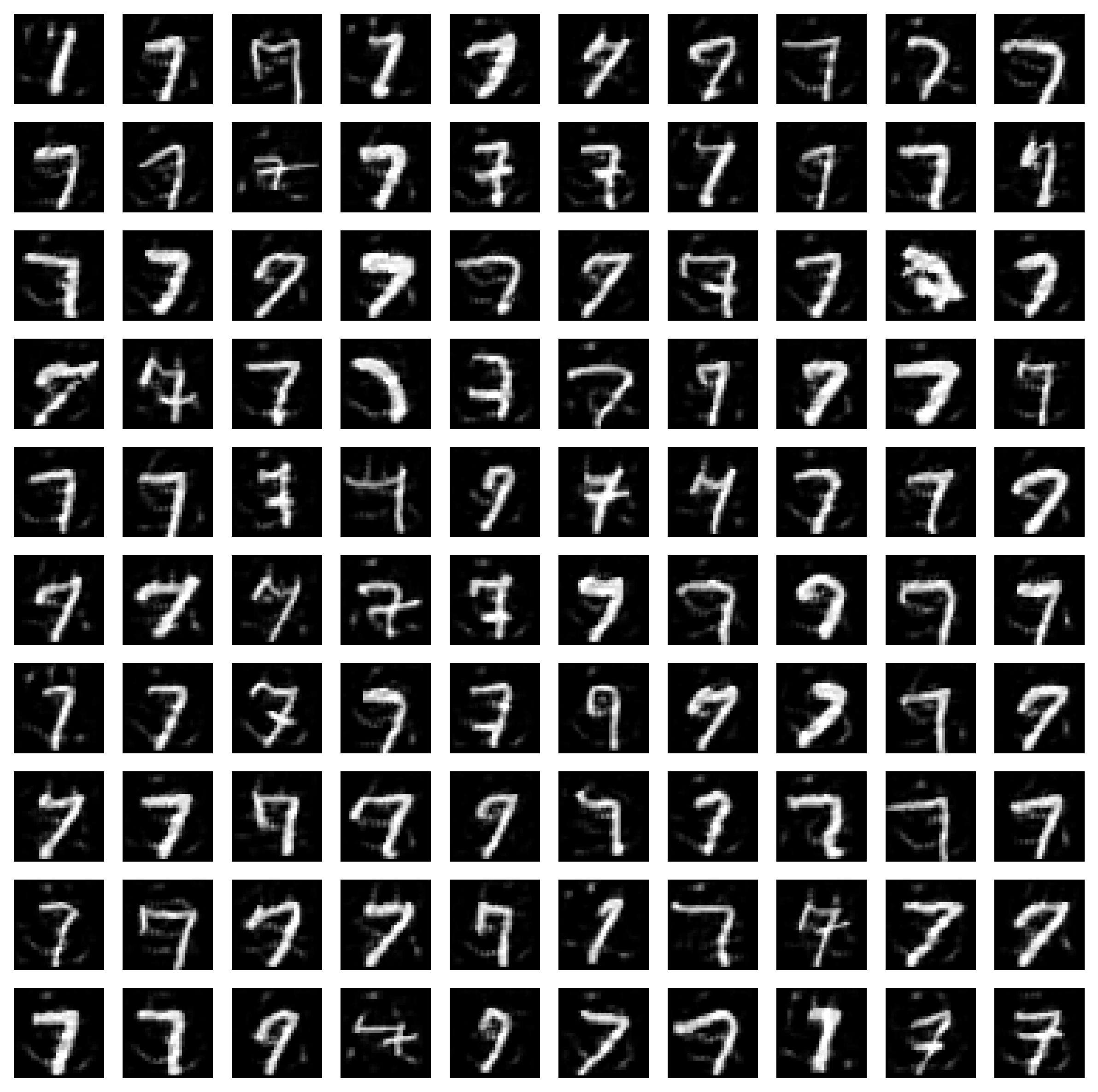}\\
\vspace{0.3in}
  \includegraphics[height = 0.45\textheight]{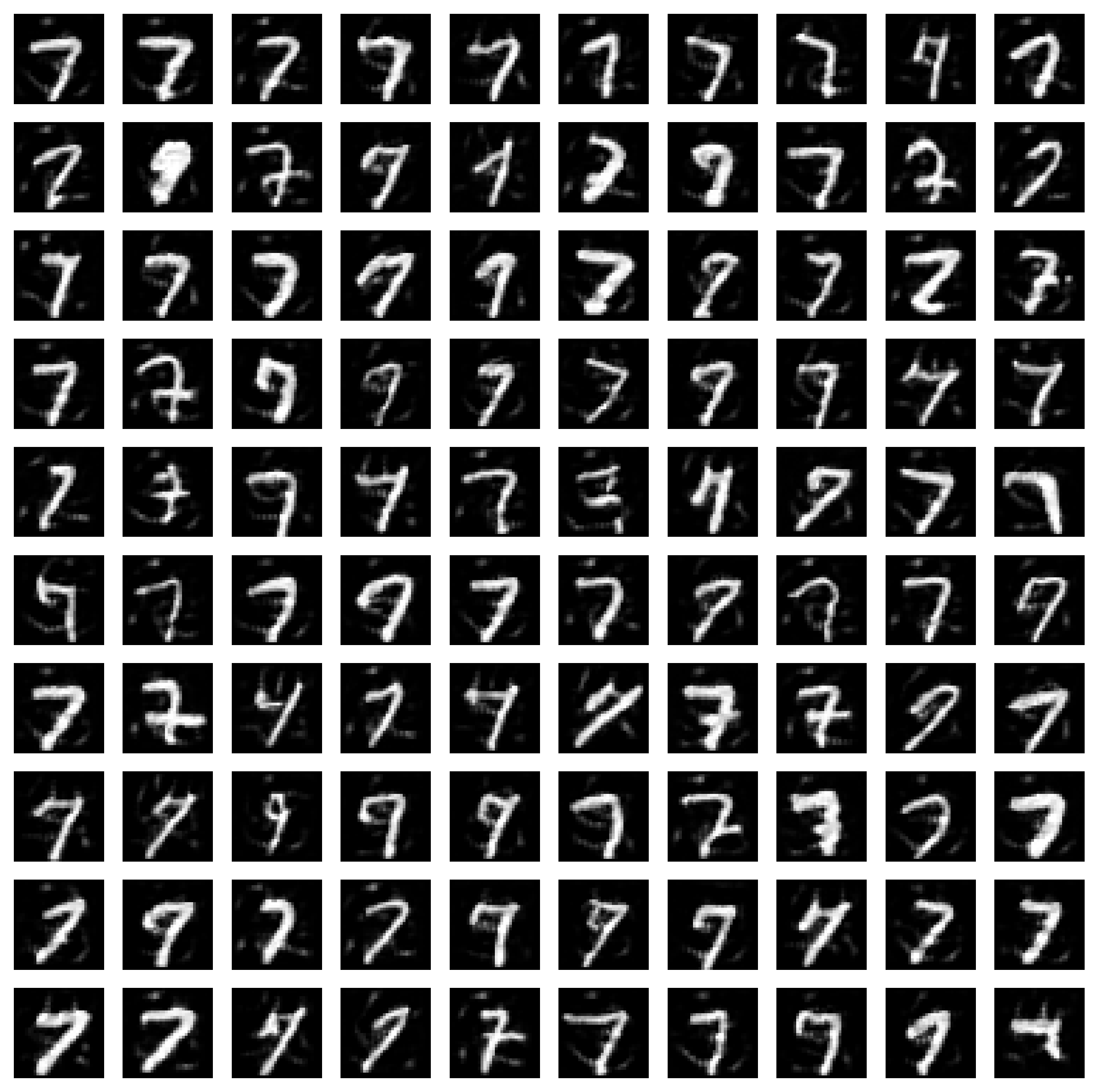} 
    \caption{Upper: 
     example of 100 images arising when the deceptive diffusion model was given the label `7'.
    Lower:
    example of 100 images arising from successful PGDL2 attacks on images that had label `7'.
    }
    \label{fig:grid7}
\end{figure}

\begin{figure}
    \centering
     \includegraphics[height = 0.45\textheight]{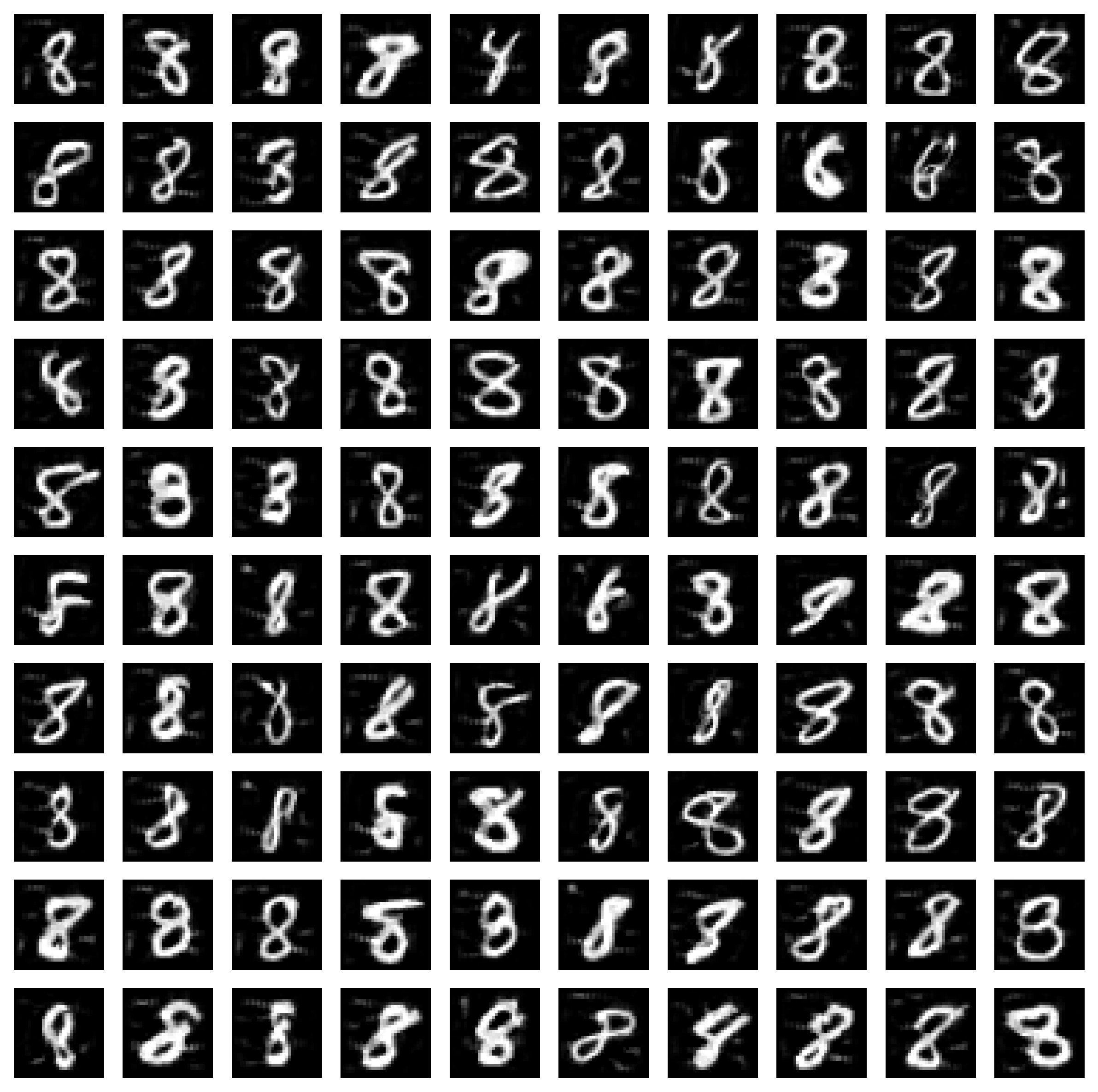}\\
\vspace{0.3in}
  \includegraphics[height = 0.45\textheight]{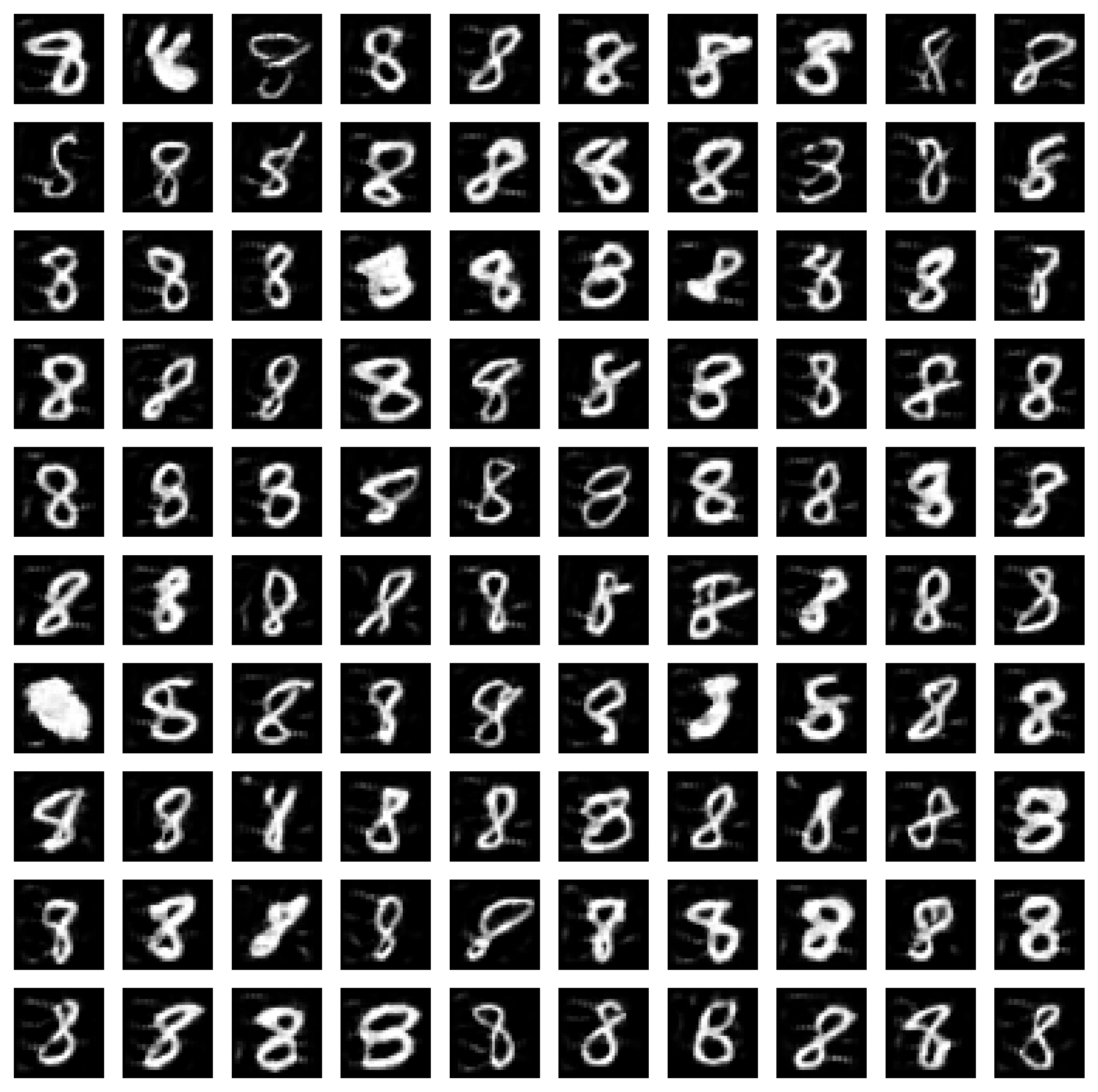} 
    \caption{Upper: 
     example of 100 images arising when the deceptive diffusion model was given the label `8'.
    Lower:
    example of 100 images arising from successful PGDL2 attacks on images that had label `8'.
    }
    \label{fig:grid8}
\end{figure}

In Figures~\ref{fig:grid0} to \ref{fig:grid8} we give analogues of 
Figure~\ref{fig:nines}
for the categories 
`0', `1',`2',\ldots,`8'.

\section{Confusion matrices for partial attacks}\label{sec:appendix_confusion}
\begin{figure}
    \centering
    \includegraphics[height = 0.6\textheight]{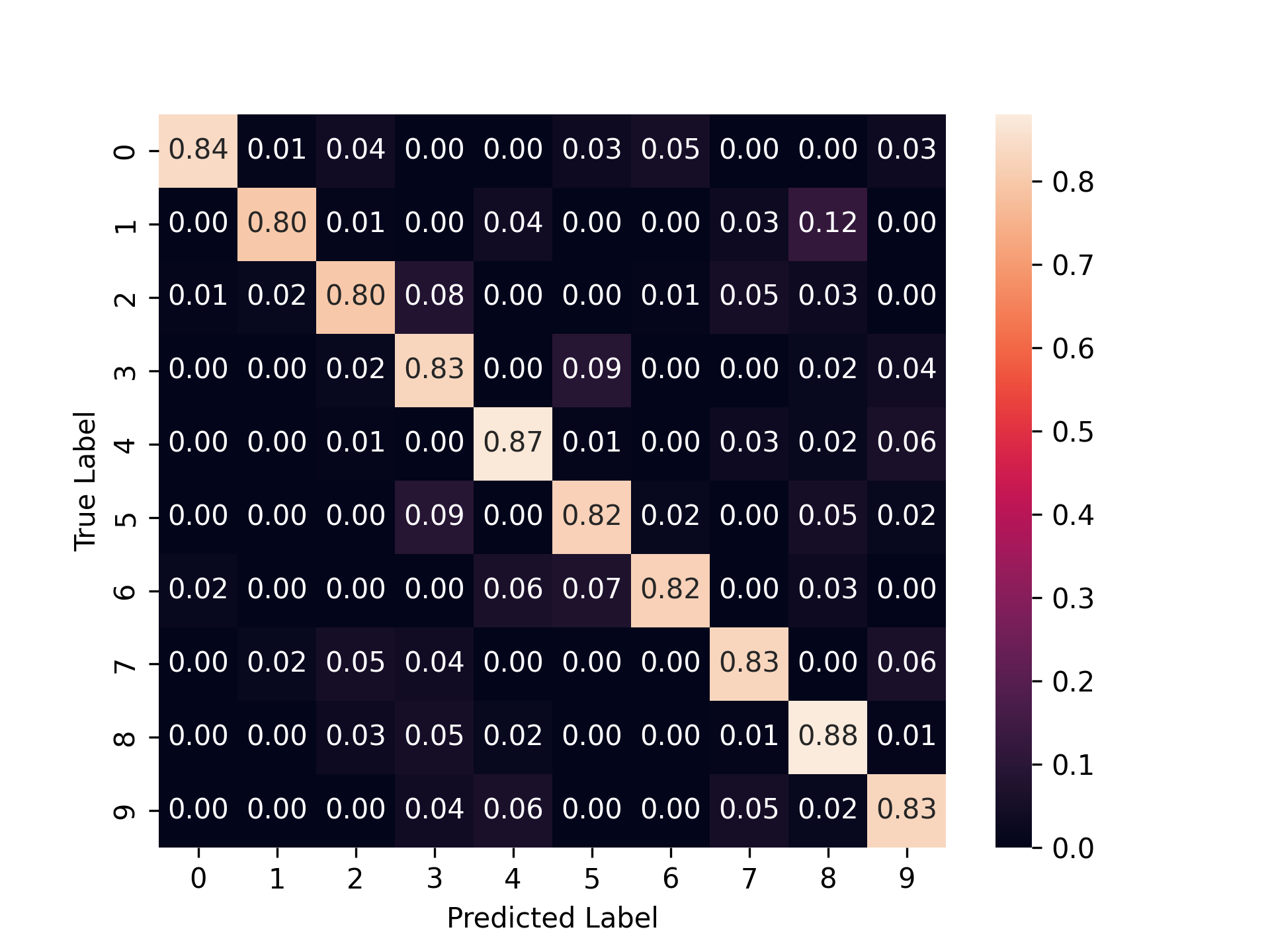} 
    \caption{Confusion matrix for the deceptive diffusion model trained with 20\% attacked data. For a given label (row) we show the frequency with which the classifier assigned each label (column) to the output. 
    Entries on the diagonal therefore correspond to 
    unsuccessful attempts to create an adversarial image.
     Overall misclassification rate is 16.8\%.
    }
    \label{fig:confusion_20}
\end{figure}

\begin{figure}
    \centering
    \includegraphics[height = 0.6\textheight]{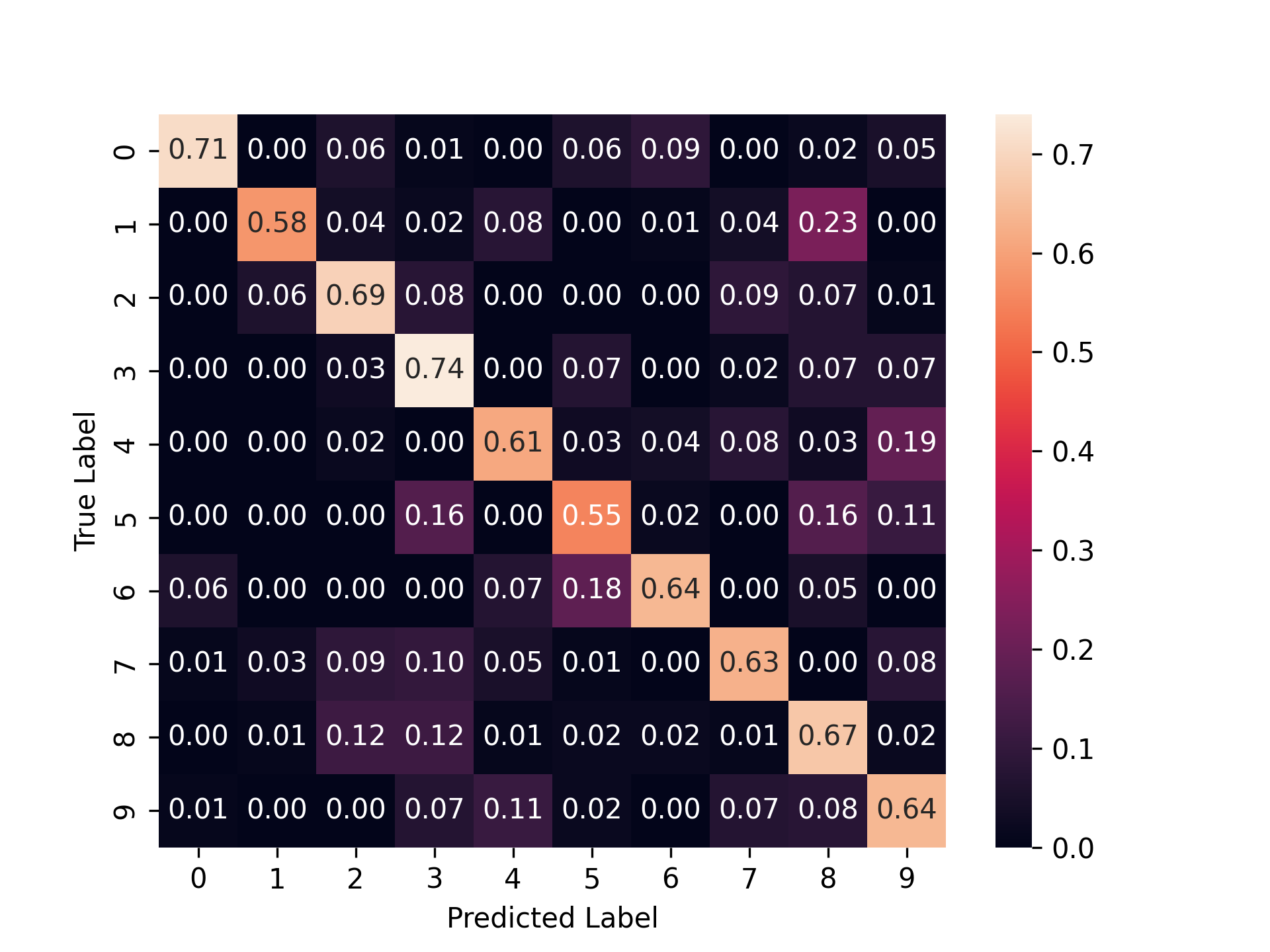} 
    \caption{Confusion matrix for the deceptive diffusion model trained with 40\% attacked data. For a given label (row) we show the frequency with which the classifier assigned each label (column) to the output. 
    Entries on the diagonal therefore correspond to 
    unsuccessful attempts to create an adversarial image.
     Overall misclassification rate is 35.4\%.
    }
    \label{fig:confusion_40}
\end{figure}

\begin{figure}
    \centering
    \includegraphics[height = 0.6\textheight]{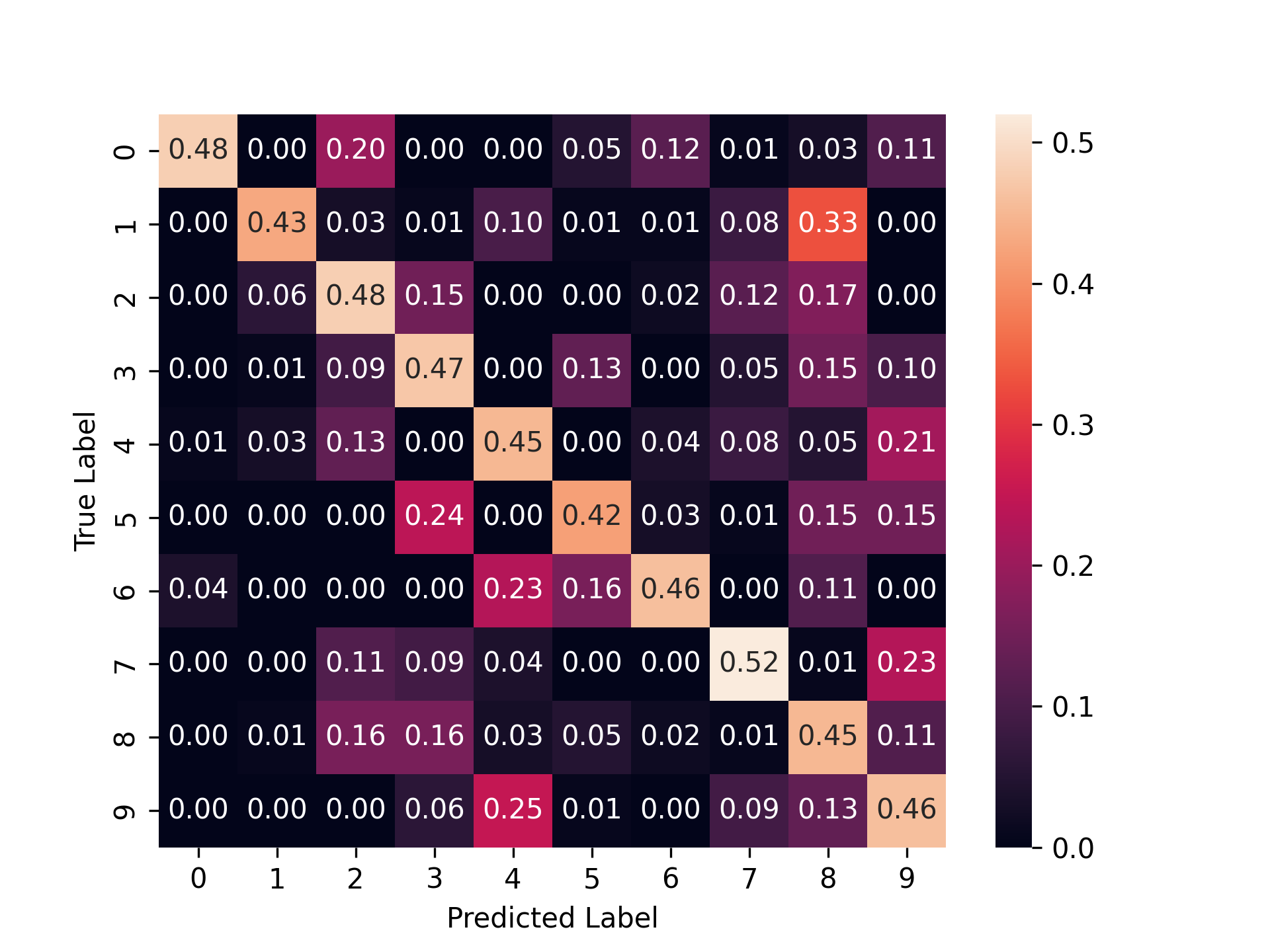} 
    \caption{Confusion matrix for the deceptive diffusion model trained with 60\% attacked data. For a given label (row) we show the frequency with which the classifier assigned each label (column) to the output. 
    Entries on the diagonal therefore correspond to 
    unsuccessful attempts to create an adversarial image.
     Overall misclassification rate is 53.8\%.
    }
    \label{fig:confusion_60}
\end{figure}

\begin{figure}
    \centering
    \includegraphics[height = 0.6\textheight]{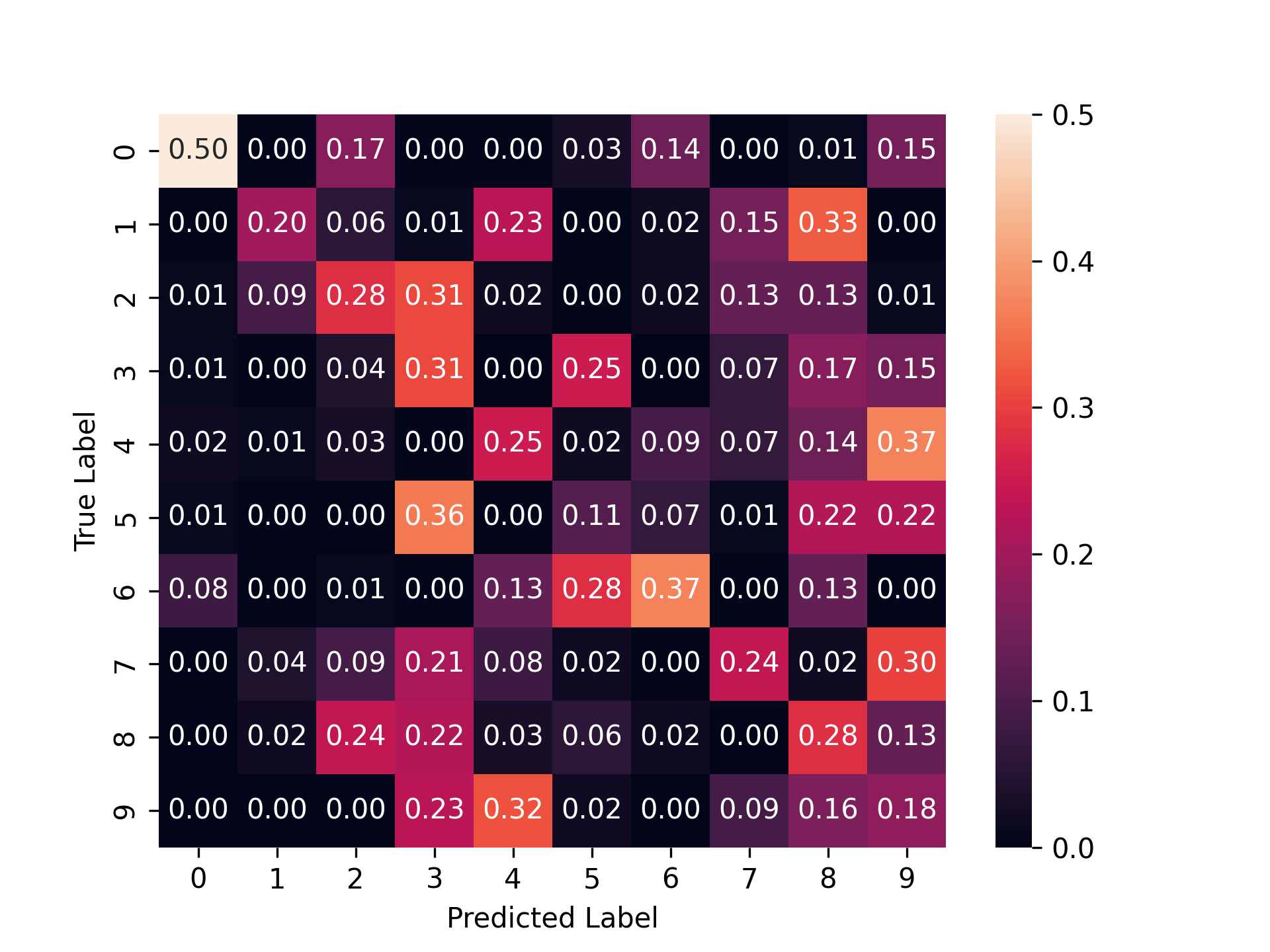} 
    \caption{Confusion matrix for the deceptive diffusion model trained with 80\% attacked data. For a given label (row) we show the frequency with which the classifier assigned each label (column) to the output. 
    Entries on the diagonal therefore correspond to 
    unsuccessful attempts to create an adversarial image.
     Overall misclassification rate is 72.8\%.
    }
    \label{fig:confusion_80}
\end{figure}

In Figures~\ref{fig:confusion_20} to \ref{fig:confusion_80} we give the confusion matrices for the deceptive diffusion model trained with $p$\% attacked data where $p\in\{20,40,60,80\}$.

\end{document}